\documentclass{article}

\usepackage{microtype}
\usepackage{graphicx}
\usepackage{booktabs} 

\usepackage{hyperref}

\usepackage[accepted]{icml2021}

\icmltitlerunning{PopSkipJump: Decision-Based Attack for Probabilistic Classifiers}

\usepackage{amsmath}
\usepackage{amsfonts}
\usepackage{bm}
\usepackage{upgreek}
\usepackage{mathtools}
\usepackage{nicefrac}

\DeclareMathOperator{\rcf}{\upphi}  
\DeclareMathOperator{\cf}{\phi}  
\DeclareMathOperator{\ar}{A\!R}  

\DeclareMathOperator{\sigmoid}{\sigma}
\DeclareMathOperator{\acq}{acq}
\DeclareMathOperator{\loss}{\mathcal{L}}  
\DeclareMathOperator{\erf}{erf}  
\DeclareMathOperator{\clip}{clip}  
\DeclareMathOperator*{\E}{\mathbb{E}}  
\DeclarePairedDelimiter{\norm}{\|}{\|_2}
\DeclarePairedDelimiter{\gennorm}{\|}{\|}  
\DeclarePairedDelimiterX{\ipd}[2]{\langle}{\rangle_2}{#1 \, , \, #2}
\DeclarePairedDelimiterX{\insideinfo}[2]{(}{)}{#1 \, \delimsize \| \, #2}
\newcommand{\info}[2]{\mathrm{I}\insideinfo{#1}{#2}}
\newcommand{\nn}{{\bm{\varphi}}}  
\newcommand{\nf}[2]{\nicefrac{#1}{#2}}
\newcommand{\p}{p}  
\newcommand{\bs}{\backslash}
\renewcommand{\det}{\mathrm{det}}  
\newcommand{\N}{\mathcal{N}}  
\newcommand{\std}{\beta}
\newcommand{\B}{\mathcal{B}}  
\newcommand{\diff}{\mathop{} \! \mathrm{d}} 

\renewcommand{\paragraph}[1]{\vspace{1ex}\textbf{#1}}

\def\eqref#1{equation~\ref{#1}}

\def\1{\bm{1}}

\def\eps{{\epsilon}}

\def\reps{\upepsilon}
\def\rdelta{\updelta}

\def\rc{{\textnormal{c}}}

\def\rg{{\textnormal{g}}}

\def\rs{{\textnormal{s}}}

\def\ry{{\textnormal{y}}}
\def\rz{{\textnormal{z}}}

\newcommand{\rvdelta}{{\bm{\updelta}}}

\def\rvg{{\mathbf{g}}}

\def\rvu{{\mathbf{i}}}

\def\rvu{{\mathbf{u}}}

\def\rvx{{\mathbf{x}}}

\newcommand{\vdelta}{{\bm{\delta}}}

\def\ve{{\bm{e}}}

\def\vg{{\bm{g}}}

\def\vu{{\bm{u}}}

\def\vx{{\bm{x}}}

\def\vz{{\bm{z}}}

\def\mI{{\bm{I}}}

\DeclareMathAlphabet{\mathsfit}{\encodingdefault}{\sfdefault}{m}{sl}
\SetMathAlphabet{\mathsfit}{bold}{\encodingdefault}{\sfdefault}{bx}{n}

\def\sD{{\mathbb{D}}}

\def\sH{{\mathbb{H}}}

\def\sK{{\mathbb{K}}}

\def\sN{{\mathbb{N}}}

\def\sS{{\mathbb{S}}}

\def\sX{{\mathbb{X}}}

\newcommand{\R}{\mathbb{R}}

\DeclareMathOperator*{\argmax}{arg\,max}

\usepackage{amsthm}
\usepackage[dvipsnames]{xcolor}
\usepackage{hyperref}
\definecolor{mydarkblue}{rgb}{0,0.08,0.45}  
\hypersetup{  
    colorlinks=true,
    linkcolor=mydarkblue,
    citecolor=mydarkblue,
    filecolor=mydarkblue,
    urlcolor=mydarkblue,}
\usepackage{url}
\usepackage[inline]{enumitem}
\usepackage[capitalise]{cleveref}
\usepackage{todonotes}
\usepackage{graphicx}
\usepackage{subcaption}
\usepackage{algorithm}
\usepackage{algorithmic}
\usepackage[super]{nth}
\usepackage{refcount}
\usepackage{placeins}

\crefname{enumi}{}{}
\creflabelformat{equation}{#2{#1}#3}

\allowdisplaybreaks
\graphicspath{{figures/}}

\newtheorem{theorem}{Theorem}  
\newtheorem{proposition}[theorem]{Proposition}
\newtheorem{lemma}[theorem]{Lemma}

\theoremstyle{remark}
\newtheorem{remark}[theorem]{Remark}

\def\ifcomments{\iffalse}
\newcommand{\cj}[1]{\ifcomments \textcolor{red}{cj: #1} \fi}
\newcommand{\ns}[1]{\ifcomments \textcolor{purple}{ns: #1} \fi}

\begin{document}

\twocolumn[
\icmltitle{PopSkipJump: Decision-Based Attack for Probabilistic Classifiers}




\begin{icmlauthorlist}
\icmlauthor{Carl-Johann Simon-Gabriel}{eth}
\icmlauthor{Noman Ahmed Sheikh}{eth}
\icmlauthor{Andreas Krause}{eth}
\end{icmlauthorlist}

\icmlaffiliation{eth}{ETH Zürich}

\icmlcorrespondingauthor{CJSG}{cjsg@ethz.ch}

\icmlkeywords{Adversarial examples, adversarial attacks, decision-based black-box attacks}

\vskip 0.3in
]



\printAffiliationsAndNotice{}  

\begin{abstract}
\looseness -1 Most current classifiers are vulnerable to adversarial examples, small input perturbations that change the classification output. Many existing attack algorithms cover various settings, from white-box to black-box classifiers, but typically assume that the answers are \emph{deterministic} and often fail when they are not.
We therefore propose a new adversarial decision-based attack specifically designed for classifiers with \emph{probabilistic} outputs.
It is based on the HopSkipJump attack by \citet{chen19hsja}, a strong and query efficient decision-based attack originally designed for deterministic classifiers.  
Our \emph{P}(robabilisticH)\emph{opSkipJump} attack adapts its amount of queries to maintain HopSkipJump's original output quality across various noise levels, while converging to its query efficiency as the noise level decreases.
We test our attack on various noise models, including state-of-the-art off-the-shelf randomized defenses, and show that they offer almost no extra robustness to decision-based attacks.
Code is available at {\small \url{https://github.com/cjsg/PopSkipJump}}.
\end{abstract}

\section{Introduction}

\cj{y-scale of plots should be l2-dist, not l2-dist divided by sqrt(d).}
Over the past decade, many state-of-the-art neural network classifiers turned out to be vulnerable to adversarial examples: small, targeted input perturbations that manipulate the classification output.
The many existing attack algorithms to create adversarial inputs cover a wide range of settings:
from white- to black-box algorithms (a.k.a.\ decision-based) over various gray-box and transfer-based settings, targeted and untargeted attacks for various kinds of data (images, text, speech, graphs); etc.
Decision-based attacks are arguably among the most general attacks, because they try not to rely on any classifier specific information, except final decisions.
Said differently, they can attack anything that can be queried often enough; in principle, even humans.
Surprisingly however, despite this generality, they typically cannot deal with noisy or probabilistic classification outputs --~a quite natural and common setting in the real world.
There could indeed be many reasons for this noise:
occasional measurement errors, classifiers' uncertainty about the answer, queries coming from different/changing classifiers, or intentional variations from a randomized adversarial defense.
Yet, as shown by \cref{tab:hsj-with-noise,apx:hsj-with-noise}, randomly changing the output label of only 1 out of 20 queries when applying SOTA decision-based attacks suffices to make them fail.

One way to deal with probabilistic outcomes would be to apply majority voting on repeated queries.
In practice, however, some input regions may need more queries than others, either because they are noisier, or because they are more important for the success of the attack.
So a naive implementation where every point gets queried equally often would require unnecessarily many queries, which is often not acceptable in real-world applications. 

\paragraph{Contributions}
We therefore propose to adapt \citet{chen19hsja}'s \emph{HopSkipJump} (HSJ) algorithm, a  query efficient, decision-based, iterative attack for \emph{deterministic} classifiers, to make it work with noisy, probabilistic outputs.
We take a model-based, Bayesian approach that, at every iteration, evaluates the local noise level (or probabilities) and uses it to optimally adapt the number of queries to match HSJ's original performance.
The result is a \emph{p}robabilistic version of H\emph{opSkipJump}, \emph{PopSkipJump} (PSJ), that 
\begin{enumerate}[label=\arabic*.,nosep]
\item \label{it:p1} outperforms majority voting on repeated queries;
\item \label{it:p2} efficiently adapts its amount of queries to maintain HSJ's original output quality at every iteration over increasing noise levels;
\item \label{it:p3} gracefully converges to HSJ's initial query efficiency when answers become increasingly deterministic;
\item \label{it:p4} works with various noise models.
\end{enumerate}
In particular, we test our attack on several recent state-of-the-art off-the-shelf randomized defenses, which all rely on some form of deterministic base model.
PSJ achieves the same performance as HSJ on the original base models, showing that these defense strategies offer no extra robustness to decision-based attacks.
Finally, most parameters of HSJ have a direct counterpart in PSJ.
So by optimizing the parameters of HSJ or PSJ in the deterministic setting we get automatic improvements in the various stochastic settings.

\paragraph{Related Literature.}
White-box attacks are the one extreme case where the attacker has \emph{full knowledge} of the classifiers' architecture and weights.
They typically use gradient information to find directions of high sensitivity to changes of the input,
e.g.,~FGSM \citep{goodfellow15explaining}, PGD \citep{madry18pgd}, Carlini\&Wagner attack \citep{carlini17towards} and DeepFool \citep{moosavi16deepfool}.
At the other extreme, decision-based attacks only assume access to the classifiers' \emph{decisions}, i.e.,\ the top-1 class assignments.
State-of-the-art decision-based attacks include the \emph{boundary attack} by \citet{brendel18boundary}, \emph{HopSkipJump} by \citet{chen19hsja} and \emph{qFool} by \citet{liu19qfool}.
In between these two extremes, many intermediate, gray-box settings have been considered in the literature.
When the architecture is at least roughly known but not the weights, one can use transfer attacks, which compute adversarial examples on a similar, substitute network in the hope that they will also fool (`transfer to') the targeted net.
\Citet{ilyas18black} consider cases where the attacker knows the top-$k$ output logits/probabilities, or the top-$k$ output ranks.
However, all these attacks were originally designed for deterministic classifiers, and typically break when answers are just slightly noisy. 
This has led several authors to propose alleged ``defenses'' using one or another form of randomization, such as neural dropout, adversarial smoothing, random cropping and resizing, etc. (details in \cref{sec:experiments}).
Shortly later however, \citet{athalye18obfuscated} managed to circumvent many of these randomized defenses \emph{in the white-box setting}, by adapting existing white-box attacks to cope with noise.
\Citet{cardelli19robustness,cardelli19statistical} studied \emph{white-box} attacks and certifications for Gaussian process classifiers and Bayesian nets.
To date however, and to the best of our knowledge, there has been no such attempt in the decision-based setting.
In particular, we do not know of any SOTA \emph{decision-based} attack that can handle noisy or randomized outputs.
Our paper closes this gap: we provide a decision-based attack that achieves the same performance on randomized defenses than SOTA decision-based attacks on the undefended nets.

\paragraph{Notations}
Vectors are bold italic ($\vx, \vdelta, \vg, \ldots$), their coordinates and 1d variables are non-bold italic ($x, \delta, g, \ldots$). 
Random vectors are bold and upright ($\rvdelta, \rvg, \rcf, \ldots$), their coordinates and 1d random variables are non-bold upright ($\rdelta, \rg, \cf, \ldots$)
When it cannot be avoided, we will also use the index-notation (e.g., $\rvdelta_i$) to designate the ($i$-th) coordinate of vector ($\rvdelta$).
$\mathrm{Ber}(p)$ denotes the Bernoulli distribution \emph{with values in $\{-1,1\}$}, returning $1$ with probability $p$.

\begin{table}[tb]
\begin{center}
\begin{tabular}{|c|c|c|c|}
\hline
\textbf{FLIP}  &\textbf{PSJ} &\textbf{HSJ}  &\textbf{HSJ x3} \\  %
\hline
$\nu=0\%$   &0.006 (1x)    &0.006 (0.90x)    &0.005 (2.70x)    \\  
$\nu=5\%$   &0.006 (1x)    &0.022 (0.74x)   &0.007 (2.35x)   \\  
$\nu=10\%$  &0.006 (1x)    &0.036 (0.73x)  &0.013 (2.27x)  \\  
\hline
\end{tabular}
\end{center}
\caption{Median size of adversarial perturbation (``border-distance'', see \cref{sec:experiments}) and relative number of model calls (in brackets) for different attacks (columns) and different noise levels (rows) on MNIST.
All attacks perform similarly on deterministic classifiers.
Randomly changing $\nu$ percent of outputs however suffices to break HopSkipJump (HSJ), even with majority voting on 3 repeated queries, whereas PopSkipJump (PSJ, our method) remains unaffected.%
}
\label{tab:hsj-with-noise}
\vspace{-.5em}
\end{table}

\section{Probabilistic Classifiers\label{sec:prob-classifiers}}

\paragraph{Definition.}
We define a probabilistic classifier as a random function (or stochastic process) $\rcf$  from a set of inputs $\sX$ to a set of $K$ classes $\sK := \{1, ..., K\}$.
Said differently, for any $\vx \in \sX$, $\rcf(\vx)$ is a random variable taking values in $\sK$.%
\footnote{Note however that, starting from \cref{sec:algo}, we will convene that $\rcf$ takes its values in $\{-1,1\}$.}
Repeated queries of $\rcf$ at $\vx$ yield i.i.d.\ copies of $\rcf(\vx)$.
\begin{remark}\label{rmk:logits}
For every probabilistic classifier $\varphi$, there exists a function of logits $\vx \mapsto \nn(\vx) := (\nn(\vx)_1, \ldots, \nn(\vx)_K) \in (\R \cup \{\pm \infty \})^K$
which (after a softmax) defines the distribution of $\rcf(\vx)$ at every point $\vx$.
Conversely, every such logit function defines a unique probabilistic classifier.
So a probabilistic classifier is nothing but an arbitrary function $\nn$ with values in $(\R \cup \{\pm \infty \})^K$ for which querying at $\vx$ means returning a random draw $\rcf(\vx)$ from the distribution defined by the logits $\nn(\vx)$.
In this paper, $\nn$ is a priori unknown.
But it can of course be recovered with arbitrary precision at any point $\vx$ by repeatedly querying $\rcf$ at $\vx$.
\end{remark}

\paragraph{Examples.}
Any deterministic classifier can be turned into a probabilistic classifier with noise level $\nu$ by swapping the original output with probability $\nu$ for another label, chosen uniformly at random among the remaining classes.
This could, e.g.,  model a noisy communication channel between the classifier and the attacker, noisy human answers, or answers that get drawn at random from a set of different classifiers.
Sometimes, noise is also injected intentionally into the classifier as a form of defense, e.g.,~by adding Gaussian noise to the inputs as in adversarial smoothing \citep{cohen19certified}, or via dropout of neural weights \citep{cardelli19statistical,feinman17detecting}, random cropping, resizing and/or compression of the inputs \cite{guo18countering}.
All these defenses yield random outputs, i.e., probabilistic classifiers.
Finally, any neural network with logit outputs $\nn(\vx) \in \R^K$ can be turned into a probabilistic classifier by sampling from its logits instead of returning the usual $\argmax_{k} \nn(\vx)_k$.
Even though perhaps not common in practice, such randomized selection is indeed used, e.g., in softmax-exploration in RL.
Moreover, as \cref{rmk:logits} shows, sampling from logits can in principle model \emph{all} the previous examples, if the neural network is given sufficient capacity to model arbitrary logit functions $\nn$, which is why we also consider this setting in our experiments (\cref{sec:experiments}).

\paragraph{Adversarial risk and accuracy.}
\looseness -1 For a given norm $\gennorm{\cdot}$, loss $\loss$ and perturbation size $\eta > 0$, we generalize the notion of $\eps$-adversarial (or robust) risk $\ar$ \citep{madry18pgd} from the deterministic to the probabilistic setting as
\begin{equation*}
    \ar(\gennorm{\cdot}, \loss, \eta, \rcf) 
    := \E_{(\rvx, \rc)} \E_{\rcf(\rvx)} \left[
        \max_{\gennorm{\vdelta} \leq \eta} 
        \loss(\rcf(\rvx + \vdelta), c)
    \right] \ .
\end{equation*}
For deterministic classifiers, the expectation is taken only over the distribution of the labeled datapoints $(\vx, c) \in \sX \times \sK$.
For probabilistic classifiers, it is also taken over the randomness of the output $\rcf(\vx)$ (at fixed input $\vx$).
We define adversarial accuracy as $1 - \ar(\loss_{0/1})$ where $\loss_{0/1}$ denotes the 0-1 loss.
In this work, $\gennorm{\cdot}$ will always be the $\ell_2$-norm $\norm{\cdot}$.

\section{The PopSkipJump Algorithm}\label{sec:algo}

From now on, let us fix the attacked image $\vx_*$ with true label $c$, and assume that the classifier $\rcf$ returns values in $\{-1,1\}$ with $1$ meaning class $c$ and $-1$ not class $c$.

\subsection{HopSkipJump algorithm for deterministic outputs\label{sec:HSJ}}

\begin{figure}
    \centering
    \includegraphics[width=\linewidth]{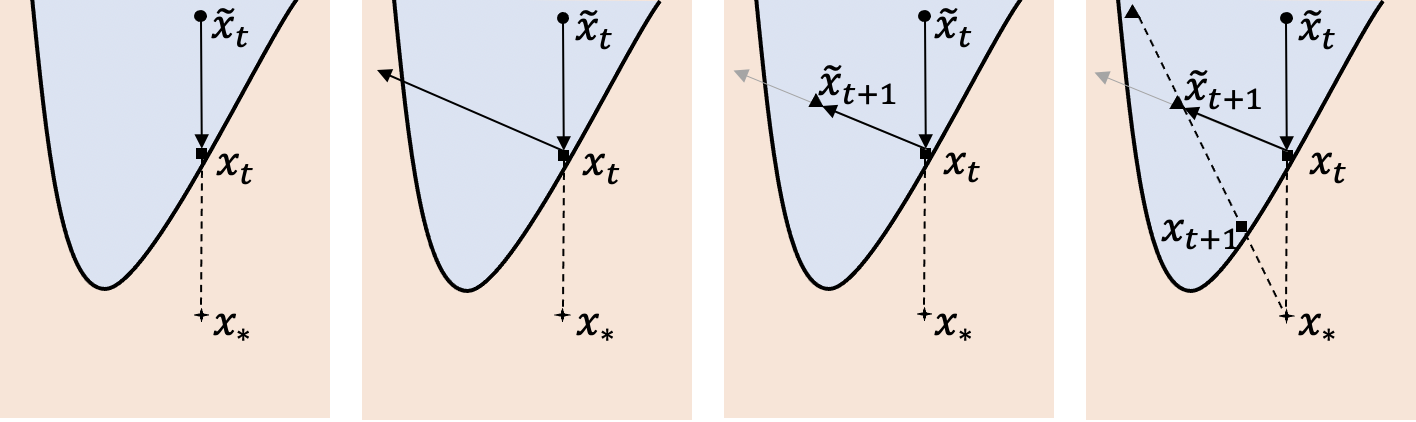}
    \caption{Original HopSkipJump algorithm for deterministic classifiers \citep{chen19hsja}.
    Illustrations correspond to steps (a), (b), (c) and (a) from the text.}
    \label{fig:hsj}
\end{figure}

Given an image $\vx_*$ with label $c$ and a neural network classifier $\nn$ which correctly assigns label $c$ to $\vx_*$, i.e., $c = \argmax_{k \in \sK} \nn(\vx_*)_k$, and let $b(\vx) := \nn(\vx)_c - \max_{k \in \sK \bs \{c\}} \nn(\vx)_k$. 
Then $b(\vx) > 0$ if $\nn$ assigns class $c$ to $\vx$, and $b(\vx) < 0$ if it does not.
Consequently, we call \emph{boundary} the set of points $\vx$ such that $b(\vx) = 0$.

Similar to the boundary attack \citep{brendel18boundary} or qFool \citep{liu19qfool}, HopSkipJump (HSJ) is a decision based attack that gradually improves an adversarial proposal $\vx_t$ over iterations $t$ by moving it along the decision boundary to get closer to the attacked image $\vx_*$.
More precisely, each iteration consists of three steps (see \cref{fig:hsj}):
\begin{enumerate}[label=(\alph*)]
    \item \emph{binary search}\label{it:bin-search} on the line between an adversarial image $\tilde{\vx}_t$ and the target $\vx_{*}$, which yields an adversarial point $\vx_t$ near the classification border.
    \item \emph{gradient estimation},\label{it:grad-est} which estimates $\vg(\vx_t) := \nabla_{\vx} b(\vx_t) / \norm{\nabla_{\vx} b(\vx_t)}$, the normal vector to the boundary at $\vx_t$, as
    \begin{equation}\label{eq:hatg}
        \hat{\vg}(\vx_t) := \frac{\vu}{\norm{\vu}}
        \ \text{with} \
        \vu := \sum_{i=1}^{n_t} \rcf(\vx_t + \vdelta^{(i)}) \vdelta^{(i)}
    \end{equation}
\looseness -1 where the $\vdelta^{(i)}$ are uniform i.i.d.\ samples from a centered sphere with radius $\delta_t$.  
We will often simply refer to $\vg$ as ``the gradient'' and to $\hat{\vg}$ as ``the gradient estimate''.
    \item\label{it:grad-step} \emph{gradient step}, a step of size $\xi_t$ in the direction
    of the gradient estimate $\hat{\vg}(\vx_t)$, yielding $\tilde{\vx}_{t+1} := \vx_t + \xi_t \hat{\vg}(\vx_t)$.
\end{enumerate}
\Citet{chen19hsja} provide various convergence results to justify their approach and fix the size of the main parameters, which are
\begin{enumerate*}[label=(\alph*)]
    \item the minimal bin size $\theta^{\det}_t = d^{-3/2} \norm{\vx_t - \vx_*}$ for stopping the binary search;
    \item the sample size $n^{\det}_t = n^{\det}_0 \sqrt{t}$ and sampling radius $\delta^{\det}_t = \theta^{\det}_t \sqrt{d}$ used to estimate the gradient;
    \item the step size $\xi^{\det}_t = \norm{\vx_t - \vx_{*}} / \sqrt t$.
\end{enumerate*}

\subsection{From HopSkipJump to PopSkipJump}

\begin{algorithm*}[tb]
   \caption{PopSkipJump}
   \label{alg:psj}
\begin{algorithmic}
    \STATE {\bfseries Input:} attacked point $\vx_*$; starting point $\tilde{\vx}_0$ from adversarial class; probabilistic classifier $\rcf$; input dim $d$;\\
    HSJ parameters: sampling sizes $n_t^{\det}$, sampling radii $\delta_t^{\det}$, min bin-sizes $\theta_t^{\det}$ and gradient step sizes $\xi_t^{\det}$.
    \FOR{$t=1$ {\bfseries to} end}
    \STATE{
        \emph{\# Compute expected cosine $C^{\det}_t$ of the HSJ gradient estimate when classifier is deterministic}\\
        $C^{\det}_t \leftarrow f(s=\infty, \eps=0, n=n_t^{\det}, \Delta=\theta_t^{\det})$ where $f = $ RHS of \cref{eq:cos}
    }
    \vspace{.5\baselineskip}
    \STATE{
        \emph{\# Do noisy bin-search  btw.\ $\tilde{\vx}_t$ and $\vx_*$ with target cos $= C^{\det}_t$ used in the stopping criterion} \\
        $\mathrm{posterior}(z, s, \eps) \leftarrow \mathrm{NoisyBinSearch}(\tilde{\vx_t}, \vx_*, C^{\det}_t)$ 
            \qquad \quad \ \emph{\# $(z,s,\eps) =$ sigmoid parameters}\\
        $(\hat{z}, \hat{s}, \hat{\eps}) \leftarrow \E_{\mathrm{posterior}}[(z,s,\eps)]$
            \quad \emph{\# Compute mean a posteriori of sigmoid parameters}\\
        $\vx_t \leftarrow \hat{z} \tilde{\vx}_t + (1 - \hat{z}) \vx_*$
            \qquad\quad\: \emph{\# Move to sigmoid center (``border'')}
    }
    \vspace{.5\baselineskip}
    \STATE{
        \emph{\# Compute query size $n_t$ for grad estimate on prob classifier to match HSJ's performance on det classifier}\\
        $n_t = \E_{z \sim \mathrm{post}}[f^{-1}(\Delta=z-\hat{z}, s=\hat{s}, \eps=\hat{\eps}, C=C^{\det}_t)]$ where $f^{-1} =$ RHS \cref{eq:n}
    }
    \vspace{.5\baselineskip}
    \STATE{
        $\hat{\vg}(\vx_t) = $ RHS of \cref{eq:hatg} \qquad \quad \quad \, \emph{\# Estimate gradient $\vg$ at point ${\vx}_t$}
    }
    \STATE{
        $\tilde{\vx}_{t+1} \leftarrow \vx_t + \xi_t \hat{\vg}(\vx_t)$ \qquad\quad\quad\, \emph{\# Make step in the estimated gradient direction}
    }
    \STATE{
        $\tilde{\vx}_{t+1} \leftarrow \vx_* + 1.5 (\vx_{t+1} - \vx_*)$ \quad \emph{\# Enlarge obtained interval $[\vx_{t+1}, \vx_*]$ to improve next bin-search}
    }
    \ENDFOR \ {\bfseries and return $\vx_t, \hat{z}, \hat{s}, \hat{\eps}$}
    \cj{highlight diffs with HSJ, add specific algo for noisy bin search}
\end{algorithmic}
\end{algorithm*}

\begin{algorithm}[tb]
   \caption{NoisyBinSearch}
   \label{alg:noisy-bin-search}
\begin{algorithmic}
    {\small
    \STATE {\bfseries Input:} attacked point $\vx_*$; $\tilde{\vx}_t$ from adv.\ class; probabilistic classifier $\rcf$ with query model $\rcf(x) \sim p_c(x)$ from \cref{eq:sigmoid}; target cos $C^{\det}_t$; prior $p(z,s,\eps)$
    \FOR{$i=1$ {\bfseries to} $\infty$}
    \STATE{
        \emph{\# Compute acquisition fct for all $x \in [\tilde{\vx}_t, \vx_*]$}\\
        $a(x) = I(\rcf(x), \rz, \rs, \reps)$ with $p(z,s,\eps)$ \& $\rcf(x) \sim p_c(x)$ 
    }
    \vspace{.3\baselineskip}
    \STATE{
        \emph{\# Sample at argmax of acquisition function}\\
        $\cf(\hat{x}) \sim p_c(\hat{x})$ where $\hat{x} = \argmax_x a(x)$
    }
    \vspace{.3\baselineskip}
    \STATE{
        \emph{\# Update prior and estimates with posterior}\\
        $p(z,s,\eps) \leftarrow p(z,s,\eps | \cf(\hat{x}))$\\
        $\hat z, \hat s, \hat{\eps} \leftarrow \E_{p(z,s,\eps}[p(z,s,\eps]$
    }
    \vspace{.3\baselineskip}
    \STATE{
        \emph{\# Use current posterior to compute query size for the next grad estimate to reach an expected cos $= C^{\det}_t$}\\
        $n_i = \E_{p(z,s,\eps)} \left[f^{-1}(z-\hat{z}, s, \eps)\right]$, $f^{-1} :=$ RHS of \cref{eq:n}
    }
    \vspace{.3\baselineskip}
    \STATE{
        \emph{\# Stop if $k$ bin-search queries spare $\leq k$ grad queries}\\
        {\bfseries if} $|n_i - n_{i-k}| \leq k$ {\bfseries : break}
    }
    \ENDFOR
    \STATE {\bfseries Output:} posterior $p(z,s,\eps)$, posterior means $(\hat z, \hat s, \hat{\eps})$.
    }
\end{algorithmic}
\vspace{-.5em}
\end{algorithm}


While HSJ is very effective on deterministic classifiers, small noise on the answers suffices to break the attack: see \cref{tab:hsj-with-noise}.
This is because, for one reason, binary search is very noise sensitive:
one wrong answer of the classifier during the binary search and $\vx_t$ can end up being non-adversarial and/or far from the classification border.
For another, even if binary search worked, the gradient estimate $\hat{\vg}$ needs more sample points $\vdelta^{(i)}$ to reach the same average performance with noise than without.

\paragraph{Overview of PopSkipJump.}
To solve these issues, our \emph{P}robabilistic H\emph{opSkipJump} attack, \emph{PopSkipJump}, replaces the binary search procedure by  \emph{(sequential) Bayesian experimental design, NoisyBinSearch}, that not only yields a point $\vx_t$, but also evaluates the noise level around that point.
We then use this evaluation of the noise to compute analytically (eq.\ \ref{eq:n}) how many sample points $n_t$ we need to get a gradient estimate $\hat{\vg}(\vx_t)$ with the same expected performance --~as measured by $\E[\cos (\hat{\vg}, \vg)]$ using \cref{eq:cos}~-- than that of the same estimator with $n^{\det}_t$ points on a deterministic classifiers.
Interestingly, when the noise level decreases, our noisy binary search procedure recovers usual binary search, and $n_t$ decreases to $n^{\det}_t$.
PopSkipJump can therefore automatically adapt to the noise level and recover the original HopSkipJump algorithm if the classifier is deterministic.
We now explain in more detail the two parts of our algorithm, noisy binary search and sample size estimation.

\subsection{Noisy bin-search via Bayesian experimental design}

\paragraph{Sigmoid assumption.}
Our noisy binary search procedure assumes that the probability $p_c(\vx)$ of the class $c$ of $\vx_*$ (the attacked image) has a sigmoidal shape along the line segment $[\vx_t, \vx_*]$.
More precisely, for $\vx = (1 - x) \vx_t + x \vx_*$ with $x \in [0,1]$, we assume that
\begin{equation}\label{eq:sigmoid}
    p_c(\vx) \equiv p_c(x) = \eps + (1 - 2\eps) \sigmoid(s (x - z) )
\end{equation}
where $\eps$ models an overall noise level, and where $\sigmoid(x) := \nf{1}{1+e^{-4 x}}$ is the usual sigmoid, rescaled to get a slope $=1$ in its center $z$ when the inverse scale parameter $s$ is equal to $1$.
This assumption is particularly well-suited for the examples discussed in \cref{sec:prob-classifiers}, such as a probabilistic classifier whose answers are sampled from a final logit layer.
This assumption is also confirmed by \cref{sec:sigmoids}, where we plot the output probabilities along the bin-search line $[\vx_t, \vx_*]$ at various iterations $t$ of an attack on two sample images $\vx_*$.
Note that when $s=\infty$, we recover the deterministic case, with or without noise on top of the deterministic output, depending on $\eps$.

\paragraph{Bayesian experimental design.}
The noisy binary search procedure follows the standard paradigm of Bayesian experimental design.
We put a (joint) prior $p^{(k)}$ on $\eps$, $z$ and $s$, query the classifier at a point $x^{(k)} \in [0,1]$ to get a random label $\rcf(\vx^{(k)})$, update our prior with the posterior distribution of $(\eps, z, s)$ and iterate over these steps for $k=0, 1, ...$ until convergence.
The stopping criterion will be discussed in \cref{sec:grad-est}.
We choose $x^{(k)}$ by maximizing a so-called \emph{acquisition} function $\acq(x | p^{(k)})$, which evaluates how ``informative'' it would be to query the classifier at point $x$ given our current prior $p^{(k)}$ on its parameters.
We tested two standard acquisition functions:
\begin{enumerate*}[label=(\roman*)]
    \item mutual information $\info{\rcf(x)}{\rs,\rz,\reps}$ between the random answer $\rcf(x)$ to a query at $x$ and the parameters $s,z,\eps$;
    \item an expected improvement approach, where we choose $x$ to minimize the expected sample size $\E_{s,t,\eps}[n_t | \rcf(x)]$ that will be required for the next gradient estimation and where $n_t$ is computed using \cref{eq:n} below.
\end{enumerate*}
Mutual information worked best, which is why we keep it as default.
We can then use the final prior/posterior to get an estimate $(\hat z, \hat s, \hat{\eps})$ of the true parameters $(z, s, \eps)$, for example with the maximum or the mean a posteriori.
We compute all involved quantities by discretizing the parameter space of $(x, z, s, \eps)$ and start with uniform priors.
See details in \cref{sec:experiments}. 

\subsection{Sample size for the gradient estimate\label{sec:grad-est}}

\paragraph{From sphere to normal distribution.} Although the original gradient estimate in the HopSkipJump attack samples the perturbations $\vdelta^{(i)}$ of the near-boundary point $\vx_t$ on a sphere, we instead sample them from a normal distribution $\N(0, \std_t \, \mI_d)$ with diagonal standard deviation $\std_t := \delta_t / d$.
This will simplify our analytical derivations and makes no difference in practice, since, in high dimensions $d$, this normal distribution is almost a uniform over the sphere with radius $\delta_t$ (in particular, $\norm{\vdelta^{(i)}} \approx \delta_t$).

\paragraph{Sample size $n_t$.}
\cj{Rephrase: general sigmoid plane and include cosine projection.}
We use the previous hypothesis to approximate the expected cosine $\E[\cos(\hat{\vg}, \vg)]$ between the gradient $\vg(\vx)$ at point $\vx$ and its estimate $\hat{\vg}(\vx)$ for a sample size $n$ as follows (justification in \cref{sec:justification-cos-eq}).
\begin{multline}\label{eq:cos}
    \E[\cos(\hat{\vg}, \vg)]
    \approx \frac{1}{\sqrt{1 + \frac{d-1}{n \alpha^2}}}
    \quad \text{where} \\
    \left \{
    \begin{array}{l}
        \alpha(\Delta, s, \std, \eps)
            := \E_{\rdelta, \ry_{\eps}}\big[
            \ry_{\eps}\big(s(\Delta + \std \rdelta)\big) \rdelta\big] \\
            \rdelta \sim \N(0, 1),
            \ry_{\eps}(x) \sim \mathrm{Ber}(\eps + (1-2\eps) \sigmoid(x))\ ,
    \end{array}
    \right .
\end{multline}
where we defined the displacement $\Delta := x-z$ between the gradient sampling center $x$ and the sigmoid center $z$ and where $\mathrm{Ber}$ denotes the Bernoulli distribution \emph{with values in $\{-1,1\}$}.
This equation is easily inverted to get the sample size $n$ as a function of the expected cosine $C$:
\begin{equation}
    \label{eq:n}
    n \approx \frac{C^2}{\alpha^2(\Delta, s, \std, \eps)} \frac{d-1}{1 - C^2}
    \ , \quad
    C := \E[\cos(\hat{\vg}, \vg)] \ .
\end{equation}
Finally, the following result shows how to compute $\alpha$ when replacing the usual sigmoid by its close approximation, a clipped linear function (proof in \cref{sec:proof-alpha-eq}).

\begin{proposition}\label{prop:alpha}
Assume that, in \cref{eq:cos}, $\sigmoid$ is the clipped linear function $\sigmoid(x) = \clip (x+\nicefrac{1}{2}, 0, 1)$.
Then
\begin{align}
    \label{eq:standard-prob-grad}
    \!\!\!\alpha(\Delta, s, \std, \eps) 
    &= \left \{
    \begin{array}{l}
        (1-2\eps) s \std \ \cdot \\
        \Big(%
        \erf\big(
            \frac{\Delta + \nf{1}{2s}}{\std \sqrt{2}}
            \big)
        - \erf\big(
            \frac{\Delta - \nf{1}{2s}}{\std \sqrt{2}}
            \big)%
        \Big)
    \end{array}
    \right .\\
    \label{eq:standard-det-grad}
    \!\!\!\alpha(\Delta, \infty, \std, \eps)
        &= (1-2\eps)\sqrt{\frac{2}{\pi}} e^{-\nicefrac{(\std \Delta)^2}{2}} \ .
\end{align} 
\end{proposition}

\begin{figure}
    \centering
    \includegraphics[width=.48\linewidth]{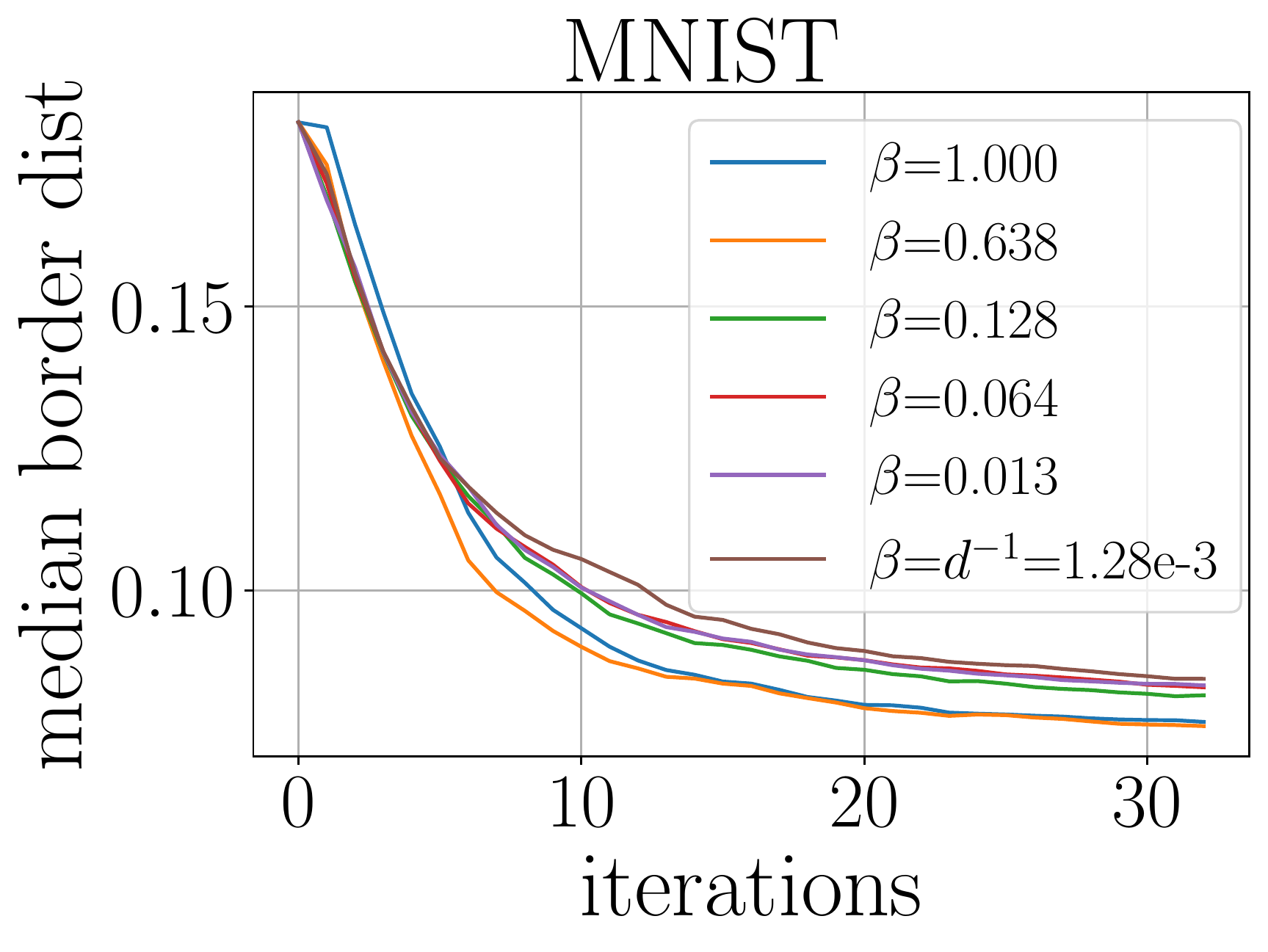}
    \includegraphics[width=.48\linewidth]{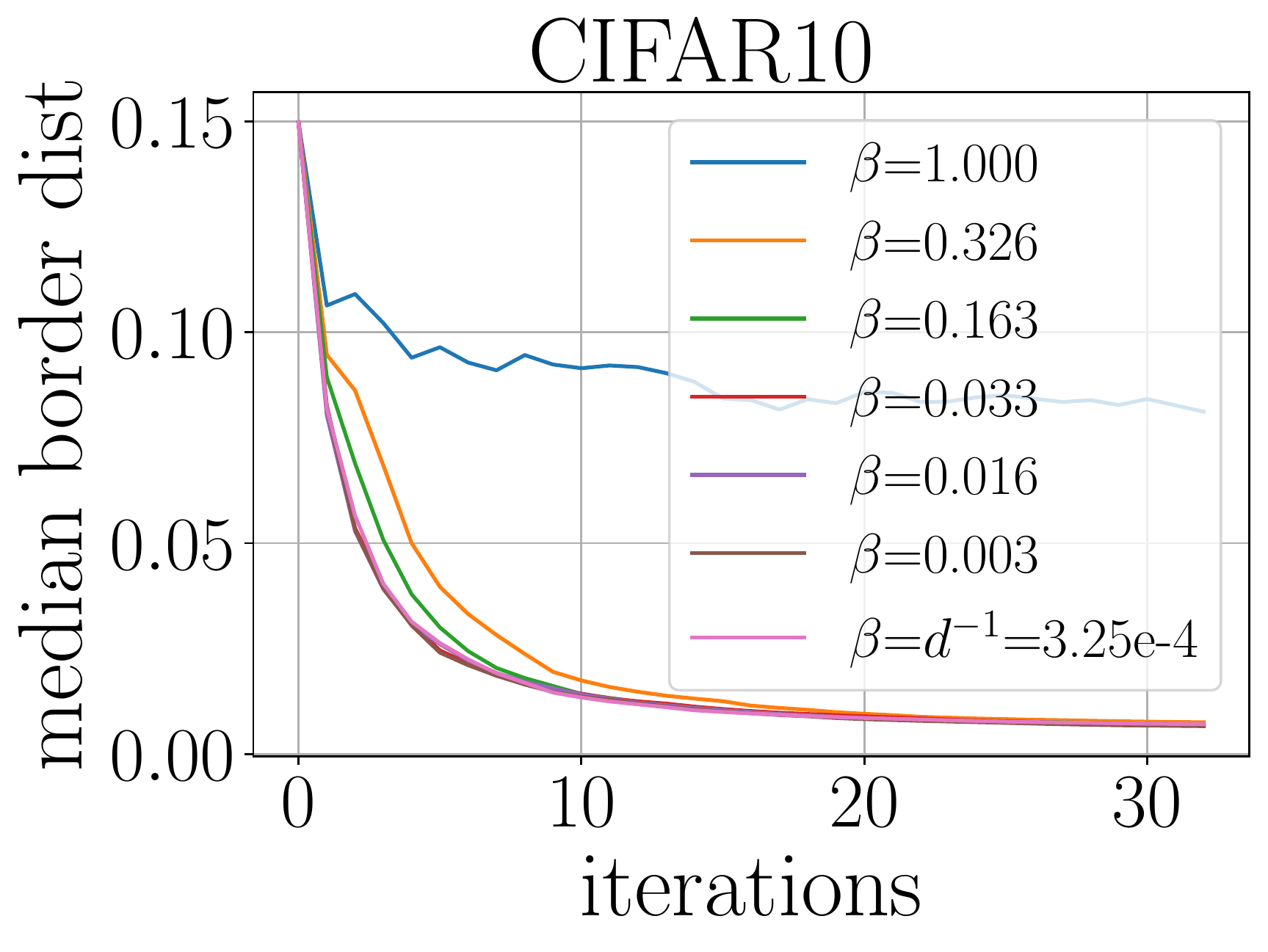}
    \caption{
    HSJ's performance on deterministic deep net classifiers is largely independent of the effective sampling radius $\delta_t = \beta \norm{\tilde{\vx_t} - \vx_*}$ of the gradient estimator.
    So we can safely increase $\beta$ (hence $\delta_t$) by several orders of magnitude, which greatly enhance PSJ's query efficiency in the noisy setting (see \cref{sec:psj-vs-hsj}) without affecting PSJ/HSJ's  output quality in the deterministic setting.
    Here we let $\beta$ range from HSJ's original choice $1/d$ to $1$, and adjusted the minimal bin-size $\theta_t$ to preserve the ratio $\delta_t / \theta_t = \sqrt d$, as in \citet{chen19hsja}, IV.C.b.\ and eq.\ 15.}
    \label{fig:various-deltas}
    \vspace{-1.5em}
\end{figure}

We now explain how to use these four formulae, together with the estimates $(\hat z, \hat s, \hat{\eps})$ from the binary search procedure, to evaluate the sample size $n_t$ that we need to get the same expected cosine value than with $n^{\det}_t$ points sampled from a deterministic classifier.
First we set $s = \infty, \eps = 0$, and $n = n^{\det}_t$ in \cref{eq:cos} and compute the expected cosine $C^{\det}_t$ on a deterministic classifier with $n^{\det}_t$ sample points;
then we apply \cref{eq:n} with our estimates $(\hat z, \hat s, \hat{\eps})$ and use the obtained value $n_t$.
(Alternatively, instead of using the point estimate $(\hat z, \hat s, \hat{\eps})$, we could also use \cref{eq:n} to compute the $\E_{z,s,\eps}[n_t]$ using the full posterior over $(t, s, \eps)$.)

\paragraph{Stopping criterion for bin-search.}
We leverage \cref{eq:n} to design a stopping criterion for the (noisy) bin-search procedure that minimizes the overall amount of queries in PSJ.
We use it to stop the binary search when one additional query there spares, on average, less than one query in the gradient estimation procedure.
Concretely, every $k$ queries (typically, $k=10$), we use our current bin-search estimate of $(z,s,\eps)$ (or the full posterior) to compute $n$ (or $\E_{z,s,\eps \sim \mathrm{posterior}}[n]$) using \cref{eq:n} and stop the binary search when the absolute difference $|n_{new} - n_{old}|$ between the new and old result is $\leq k$.
The idea is that, the better we estimate the center $z$ of the sigmoid, the closer $x$ (center of $\hat{\vg}$) will be to $z$. 
This in turn will reduce the number of queries required for $\hat{\vg}$ to reach a certain expected cosine.
(To see that, notice for example that \cref{eq:n} decreases with $|\Delta|$.)
Since a query tends to yield more information about the position of $z$ at the beginning of the bin-search procedure than later on, $|n_{new} - n_{old}|$ tends to decrease with the amount of bin-search queries.
The order of magnitude of $|\Delta|$ when meeting the stopping criterion
depends on the shape parameter $s$ and noise level $\eps$ of the underlying sigmoid.
For a deterministic classifier ($s = \infty, \eps = 0$), it must be at least of the order of $\beta$, the standard deviation of the samples $\vdelta^{(i)}$ in $\hat{\vg}$ (see eq.\ \ref{eq:hatg}):
otherwise, all points $x+\delta^{(i)}$ would belong to the same class and yield no information about the border.
(See also IV.C.a.\ in \citealt{chen19hsja}.)
But if $\beta \ll 1/s$, the characteristic size of the linear part of the sigmoid, then $|\Delta| \geq \beta$ is acceptable, as long as it is $\leq 1/s$.
\cj{Add illustration?}
Our stopping criterion provides a natural and systematic way to trade off these considerations.

\begin{figure*}[tb]
    \includegraphics[width=\linewidth]{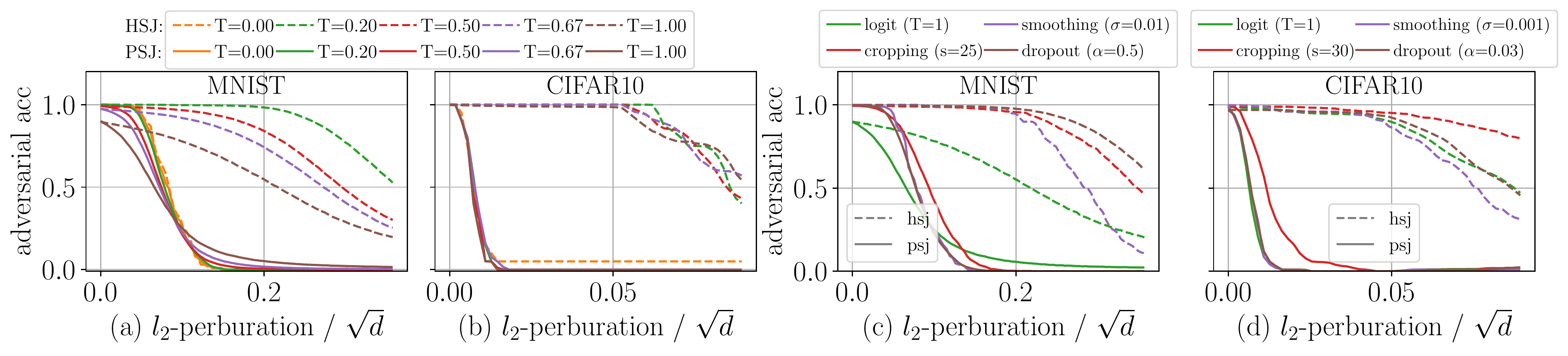}
    \caption{Adversarial accuracy versus attack size $\eta$ for PSJ and HSJ for a fixed noise model (logit-sampling) and various noise levels (temperatures $T$) in Figs.\ (a) \&(b); and for various noise models and fixed, high noise levels in Figs.\ (c) \& (d).
    The curves obtained with PSJ are well below their HSJ counterparts in all noisy settings and both curves coincide in the deterministic case (Figs.\ a \& b, T=0.).
    This illustrates the clear superiority of PSJ over HSJ.
    Note that, even though all classifiers use a same underlying base classifier, their PSJ curves do not coincide (even when $\eta=0$, i.e., for usual accuracy).
    This confirms that adversarial accuracy is ill-suited for comparisons between different noise levels and that the median border distance should be preferred, as in \cref{fig:psj-vs-noise-model,fig:psj-vs-noise-level}.
    See paragraph ``Adversarial accuracy'' and \cref{rmk:aa}. 
    \label{fig:adv-acc-vs-noise-levels}}
    \vspace{-1em}
\end{figure*}

\subsection{PopSkipJump versus HopSkipJump\label{sec:psj-vs-hsj}}

Here we discuss additional small differences between HSJ and PSJ, besides the obvious ones that we already mentioned --~binary search, its stopping criterion, and the sample size for the gradient estimate.

\paragraph{Gradients: variance reduction and size of $\delta_t$.\label{sec:delta_size}}
The authors of HSJ propose a procedure to slightly reduce the variance of the gradient estimate (Sec. III.C.c), which we do not use here.
Moreover, they use $\delta_t = \norm{\tilde{\vx_t} - \vx_*} / d$, whereas we use $\delta_t = \sqrt{d}\norm{\tilde{\vx_t} - \vx_*} / 100$.
The reason is that, whenever $s < \infty$, smaller $\delta_t$ yield noisier answers which  increases the queries needed both in the gradient estimation and in the bin-search. 
Given that the logits of deep nets typically have shape parameters $s \approx 1$, HSJ's original choice would require prohibitively many samples.
To reduce noise, the larger the radius $\delta_t$, the better.
In practice however, the size of $\delta_t$ is limited by the curvature of the border and by the validity range of the sigmoidal assumption (\cref{eq:sigmoid}).
To trade of these considerations, we evaluated the empirical performance of HSJ (on deterministic classifiers) with various choices of $\delta_t$ and chose one of the largest for which results did not differ significantly from the original ones.
See \cref{fig:various-deltas}.
A more theoretically grounded approach that would evaluate the curvature is left for future work.

\paragraph{No geometric progression on $\xi_t$.}
For HSJ, it is crucial that $\tilde{\vx_t}$ be on the adversarial side of the border.
Therefore, it always tests whether the point $\tilde{\vx_t} := \vx_{t-1} + \xi_t \hat{g}(\vx_{t-1})$ is indeed adversarial.
If not, it divides $\xi_t$ by 2 and tests again.
Since by design $\vx_{t-1}$ is adversarial, this ``geometric progression'' procedure is bound to converge.
In the probabilistic case, however, testing if a point is adversarial can be expensive, and is not needed since, on the one hand, the noisy bin-search procedure can estimate the sigmoid's parameters even if $z$ is outside of $[\tilde{\vx_t}, \vx_*]$;
and on the other, we are less interested in the point $\vx_t$ and the point value $\p_c(\vx_t)$ than in the global \emph{direction} from $\vx_*$ to $\vx_t$.
\cj{see \cref{sec:prob-classifiers}?}
We therefore use $\xi_t$ as is, without geometric progression.

\paragraph{Enlarging bin-search interval $[\tilde{\vx_t}, \vx_*]$.}
It is easier for the noisy bin-search procedure to estimate the sigmoid parameters if it can sample from both sides of the sigmoid center $z$.
In practice however, we noticed that after a few iterations $t$, the point $\tilde{\vx}_t$ tends to be very close to $z$.
We therefore increase the size of the sampling interval, from $[\tilde{\vx_t}, \vx_*]$ to $[\vx_* + 1.5 (\tilde{\vx_t}-\vx_*), \vx_*]$, which performed much better.

\section{Experiments}\label{sec:experiments}

\begin{figure*}[tb]
    \vspace{-.91em}
    \centering
    \includegraphics[width=\linewidth]{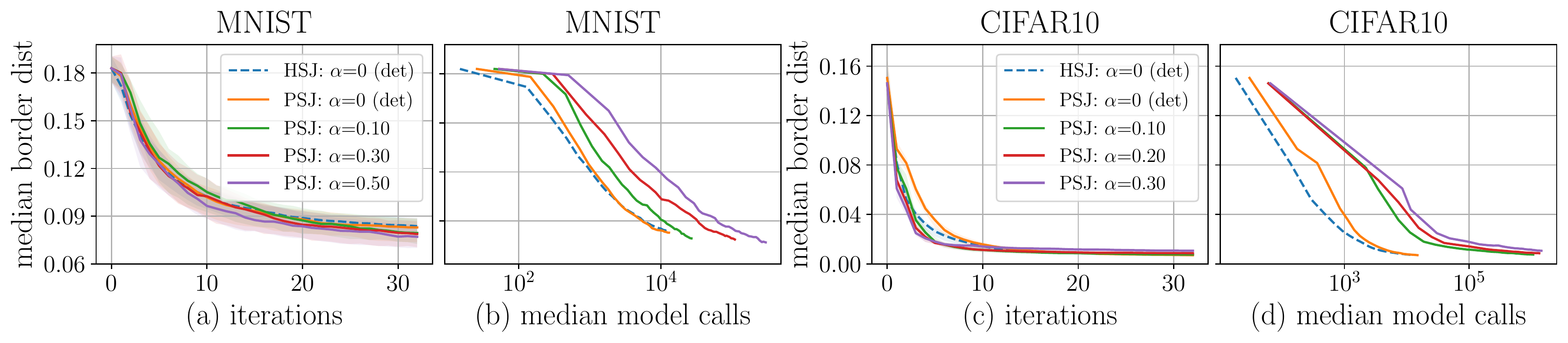}
    \caption[PSJ to HSJ]{PopSkipJump's performance (lower is better) for a fixed noise model (dropout) and various noise levels (dropout rate $\alpha$).
    Performance is shown as a function of the number of algorithm iterations (a \& c) and model queries (b \& d).
    Shaded areas depict the \nth{40} to \nth{60} percentiles.
    Plots (a) \& (c) illustrate property \cref{it:p2}:
    the \emph{per iteration} performance of PSJ is largely independent of the noise level (here, the dropout rate) and is on par with the performance achieved by HSJ on the deterministic base classifier.
    Plots (b) \& (d) illustrate property \cref{it:p1}:
    when the noise level (dropout rate) decreases and the classifier becomes increasingly deterministic, the PSJ curves converge to the limiting HSJ curve, i.e., the \emph{per query} performance of PSJ converges to that of HSJ.
    See \cref{fig:psj-vs-noise-level-complete} in appendix for similar curves, but with other noise models.
    \label{fig:psj-vs-noise-level}}
\end{figure*}

\begin{figure*}[tb]
    \centering
    \includegraphics[width=\linewidth]{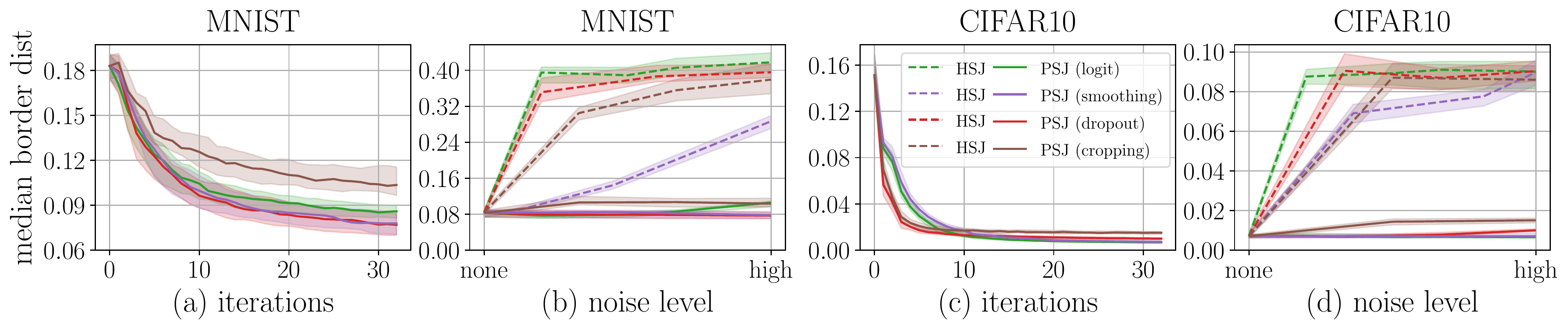}
    \caption[PSJ to HSJ]{PopSkipJump's performance (lower is better) for various noise models.
    Plots {(a) \& (c)}: Performance as a function of the number of algorithm iterations, when using, for each noise model, the highest noise levels considered in \cref{fig:psj-vs-noise-level} (see also \cref{fig:psj-vs-noise-level-complete} in appendix).
    All curves are very similar, showing that PSJ is largely invariant to the specific type of randomness used, even at high noise levels.
    Plots {(b) \& (d)}: Performance after 32 algorithm iterations (right-most part of curves in a \& c) of PSJ and HSJ as a function of the noise level.
    While small noise levels suffice to break HSJ (large border-distances at end of attack), PSJ's performance stays almost constant accross all noise levels and noise models.
    \label{fig:psj-vs-noise-model}}
    \vspace{-.91em}
\end{figure*}

\begin{figure}[tb]
    \centering
    \includegraphics[width=.48\linewidth]{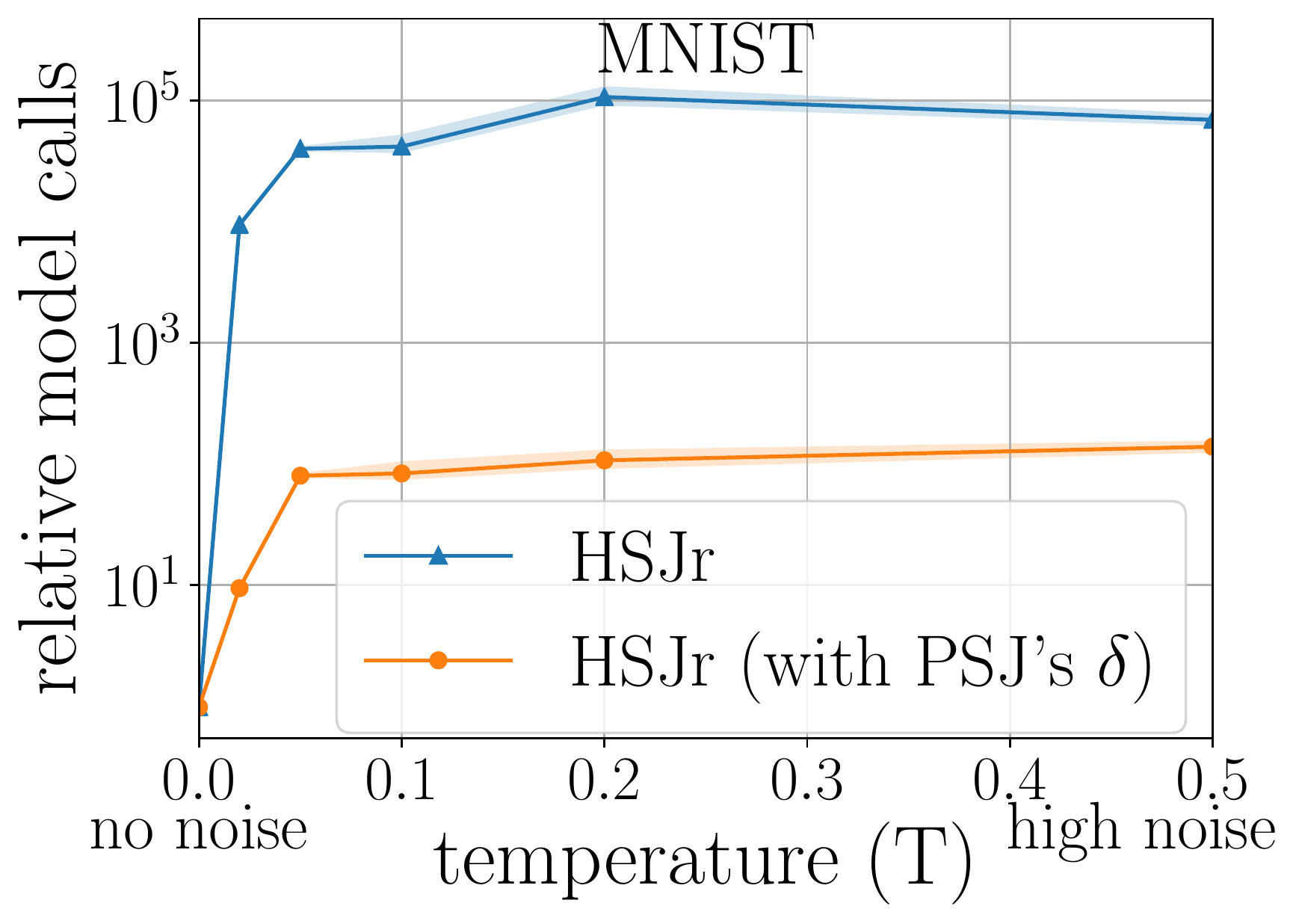}
    \includegraphics[width=.48\linewidth]{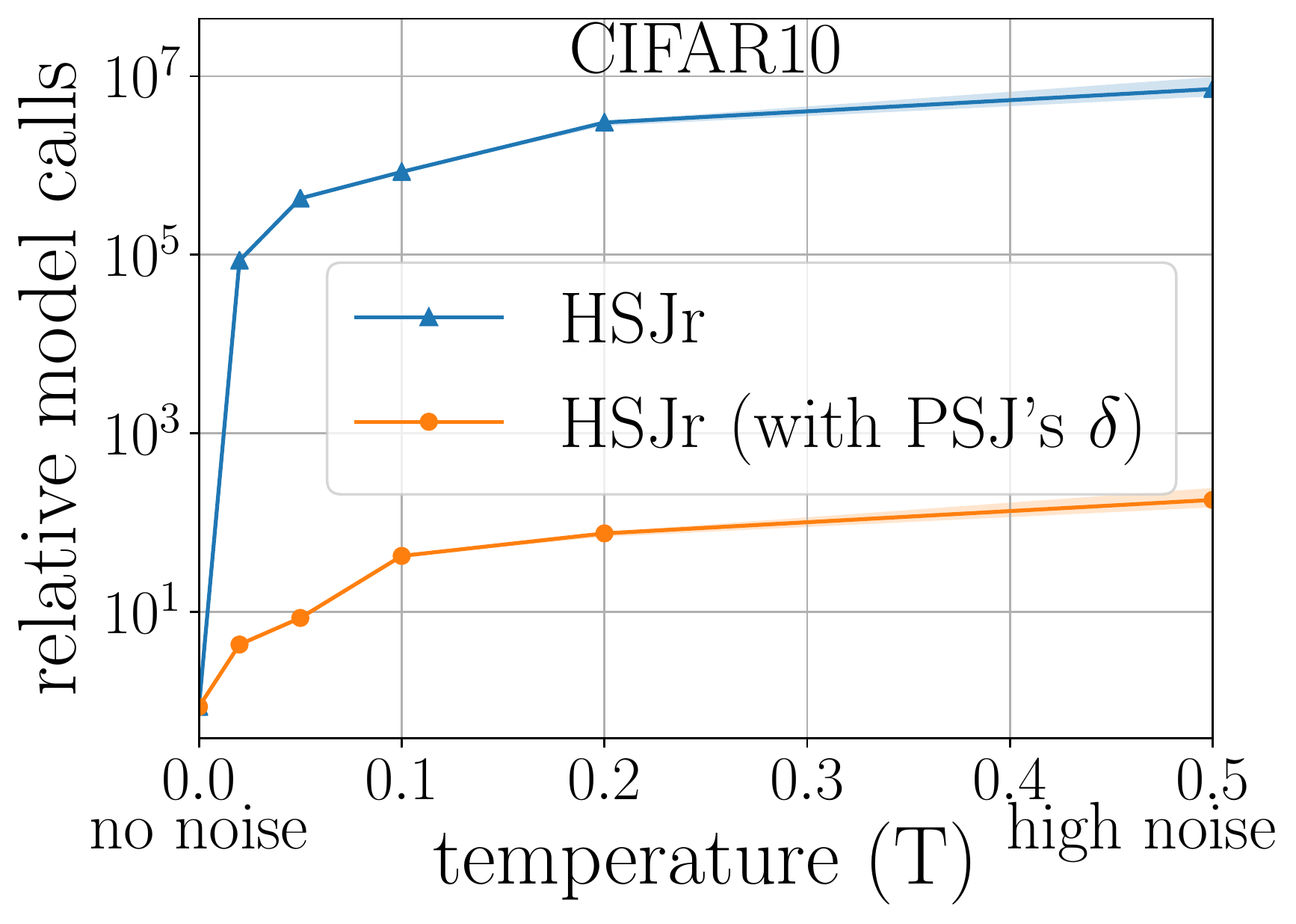}
    \caption{Ratio $N_{\mathrm{HSJr}} / N_{\mathrm{PSJ}}$ as a function of noise level, where $N_{\mathrm{HSJr}}$ is the total amount of queries needed by HSJ-with-repeated-queries to match the performance of PSJ with $N_{\mathrm{PSJ}}$ queries.
    HSJr needs several orders of magnitude more queries than PSJ, even when the classifier becomes increasingly deterministic, and even when using PSJ's larger sampling radius $\delta$ for the gradient estimator (see paragraph ``PSJ outperforms HSJr'').
    }
    \label{fig:psj-hsjr}
    \vspace{-1em}
\end{figure}

The goal of our experiments is to verify points \cref{it:p1,it:p2,it:p3,it:p4} from the introduction.
That is, we want to show that, contrary to the existing decision-based attacks, the performance of PSJ is largely independent of the strength and type of randomness considered, i.e., of both the \emph{noise level} and the \emph{noise model}.
At every iteration, PSJ adjusts its amount of queries to keep HSJ's original output quality, and is almost as query efficient as HSJ on near-deterministic classifiers.
To show all this, we apply PSJ (and other attacks) to a deterministic \emph{base classifier} whose outputs we randomize by injecting an adjustable amount of randomness.
We test various randomization methods, i.e., noise models, described below, including several randomized defenses proposed at the ICLR'18 and ICML'19 conferences.
\Cref{fig:psj-vs-noise-level,fig:psj-vs-noise-model} summarize our main results.

\begin{remark}
Although we do compare PSJ to SOTA decision-based attacks, with or without repeated queries, \emph{we do not compare PSJ to any decision-based attack specifically designed for probabilistic classifiers because, to the best of our knowledge, there is not any}.%
\footnote{The decision-based attack by \citet{ilyas18black} for deterministic classifiers may still work to some extent with randomized outputs, but it is less effective than HSJ on deterministic classifiers \citep{chen19hsja}.
Since we will show that, despite the noise, PSJ stays on par with HSJ, there is no need for further comparisons.}
There are however some \emph{white-box} attacks \citep[e.g.,][]{athalye18obfuscated,cardelli19robustness,cardelli19statistical} that can deal with some specific noise models considered in this paper (see below).
\end{remark}

\paragraph{Noise models and randomized defenses.}
For a given deterministic classifier $\nn$, we consider the following randomization schemes.
\begin{itemize}[nosep, wide, labelwidth=!,labelindent=0pt]
\item \emph{logit sampling}: divide the output logits $\nn(\vx)$ by a temperature parameter $T$ to get $\nn_T(\vx) := \nn(\vx) / T$ and sample from the new logits $\nn_T(\vx)$.
By changing $T$ we can smoothly interpolate between the deterministic classifier ($T \to 0$) and sampling from the original logits ($T=1$).

\item \emph{dropout}: apply dropout with a uniform dropout rate $\alpha \in [0,1] $\citep{srivastava14dropout}.
Taking $\alpha=0$ yields the deterministic base classifier; increasing $\alpha$ increases the randomness.
Dropout and its variants have been proposed as adversarial defenses, e.g.,\ in \citet{cardelli19statistical,feinman17detecting}.
Note that a network with dropout can be interpreted as a form of Bayesian neural net \citep{gal16dropout}.
As such, sampling from it can be understood as sampling from an ensemble of nets.

\item \emph{adversarial smoothing}: add centered Gaussian noise with standard deviation $\sigma$ to every input before passing it to the classifier.
Taking $\sigma=0$ yields the original base classifier.
\Citet{cohen19certified} proposed majority voting over several such queries as an off-the-shelf adversarial robustification.  

\item \emph{random cropping \& resizing}: randomly crop and resize every input image before passing it to the classifier.
Changing the cropping size allows to interpolate between the deterministic setting (no cropping) and more noise.
This method and a variant were proposed by \citet{guo18countering} and \citet{xie18mitigating} as adversarial defenses.
\end{itemize}

We ran all experiments on the MNIST \citep{mnist} and CIFAR10 \citep{cifar} image datasets.
Since, at high noise levels, the attack may need a million queries, it could take a minute per attack on a GeForce GTX 1080 for MNIST and a few minutes for CIFAR10 (larger net; see \cref{sec:complexity} for a time and complexity analysis and acceleration tricks.)
We therefore restricted all experiments to a same random subset of 500 images of the MNIST and CIFAR10 test sets respectively, where we kept only images that were labeled correctly with probability $\geq .75$ when using the cropping noise model with $s=22$.
On CIFAR10 we use a DenseNet-121 and on MNIST a CNN with architecture `conv2d(1, 10, 5), conv2d(10, 20, 5), dropout2d, linear(320, 50), linear(50,10)`.
In all plots, shaded areas mark the \nth{40} to \nth{60} percentiles.
To simplify the comparison across datasets (cf.\ Eq.\ 3 in \citealt{simon19advers}), we divide all $\ell_2$-distances by $\sqrt{d}$, where the input dimension $d$ is $27\times27$ for MNIST and $3\times32\times32$ for CIFAR-10.
Code is available at \url{https://github.com/cjsg/PopSkipJump}.

\paragraph{Adversarial accuracy (AA).}
\cref{fig:adv-acc-vs-noise-levels} plots adversarial accuracy as a function of the attack size $\eta$ for various noise levels and a fixed noise model using logit sampling (Figs.\ a \& b), and for various noise models at fixed, high noise levels (Figs.\ c \& d).
The accuracy curves obtained with PSJ are well below their HSJ counterparts in all noisy settings and both curves coincide in the deterministic case (Figs.\ a \& b, T=0).
This illustrates the clear superiority of PSJ over HSJ.
However, despite its standard use in the deterministic setting and its straight-forward generalization to probabilistic classifiers, AA is ill-suited for comparing performances between different noise models and noise levels.
An easy way to see this is to notice that the AA curves do not even coincide at $\eta=0$, even though the value at that point is just standard accuracy and does not depend on the attack algorithm.
Instead, we will now introduce the (median) \emph{border distance}, which generalizes the usual ``median $\ell_2$-distance of adversarial examples'' to the probabilistic setting, and which is better suited for comparisons across noise models and noise levels.

\begin{remark}\label{rmk:aa}
A deeper reason why AA is ill-suited for the comparison between noise levels is the following.
From any probabilistic classifier one can define the deterministic classifier obtained by returning, at every point, the majority vote over an infinite amount of repeated queries at that point.
This deterministic classifier is ``canonically'' associated to the probabilistic one in the sense that it defines the same classification boundaries.
Naturally, any metric that compares an attack's performance across various noise levels should be invariant to this canonical transformation.
Concretely, it means that a set of adversarial candidates $\{\vx\}$ should get the same score on a probabilistic classifier than on its deterministic counterpart.
The border distance defined below satisfies this property; AA does not.
\end{remark}

\vspace{-.5em}
\paragraph{Performance metric: border distance.}
In the deterministic case, the border distance is essentially the $\ell_2$-distance of the proposed adversarial examples to the original image.
In the probabilistic case, however, an attack like PSJ may return points that are close to the boundary, but actually lie on the wrong side, because the underlying (typically unknown) logit of the true class is only marginally greater than the logit of the adversarial one.
So, to ensure that we only measure distances to true adversarials and  \emph{for the purpose of evaluation only}, we will assume white-box access to the true underlying logits and then project all outputs $\vx$ to the closest boundary point that lies on the line $(\vx_*, \vx)$, i.e., the closest point $\vx'$ where the true and adversarial class have same probability.
We define the \emph{border-distance of $\vx$ to $\vx_*$} as the $\ell_2$-distance $\norm{\vx'-\vx_*}$ (re-scaled by $ 1 / \sqrt d$).
Note that for this evaluation metric, what matters is not so much the output point $\vx$ than finding an output-\emph{direction} $(\vx_*, \vx)$ of steep(est) descent for the underlying output  probabilities.


\paragraph{PSJ is invariant to noise level and noise model.} 
\Cref{fig:psj-vs-noise-level} fixes the noise model (dropout) and compares PSJ's performance at various noise levels (dropout rate $\alpha$).
(Similar curves for the other noise models can be found in appendix, \cref{fig:psj-vs-noise-level-complete}.)
\Cref{fig:psj-vs-noise-model} instead studies PSJ's performance on various noise models.
More precisely, \cref{fig:psj-vs-noise-level} shows the median border-distance at various noise levels (dropout rates $\alpha$) as a function of PSJ iterations (a \& c) and as a function of the median number of model queries obtained after each iteration (b \& d).
Shaded areas show the \nth{40} and \nth{60} percentiles of border-distances.
Figs.~(a) \&~(c) illustrate point \cref{it:p2} from the introduction:
the \emph{per iteration} performance of PSJ is largely independent of the noise level and on par with HSJ's performance on the deterministic base classifier.
This suggests that PSJ adapts its amount of queries \emph{optimally} to the noise level:
just enough to match HSJ's deterministic performance, and not more.
Figs.~(b) \& (d) illustrate point \cref{it:p3}:
when the noise level decreases and the classifier becomes increasingly deterministic, the \emph{per query} performance of PSJ converges to that of HSJ, in the sense that the PSJ curves become more and more similar to the limiting HSJ curve.
Note that the log-scale of the x-axis can amplify small, irrelevant difference at the very beginning of the attack.
\Cref{fig:psj-vs-noise-model} show that the performance of PSJ is largely invariant to the different noise models considered here.
Figs.~\ref{fig:psj-vs-noise-model} (b) \& (d) also confirm that, contrary to HSJ that fails even with small noise, PSJ is largely invariant to changing noise levels and noise models.

\paragraph{PSJ outperforms HSJ-with-repeated-queries.}
Let HSJ-$r$ be the HSJ attack with majority voting on $r$ repeated queries at every point.
\Cref{fig:psj-hsjr} studies how many total queries HSJ-$r$ requires to match the performance of PSJ at various noise levels with logit-sampling.
It reports the ratio $N_{\mathrm{HJSr}} / N_{\mathrm{PSJr}}$ of total amount of queries.
Concretely, for logit sampling with a given temperature $T$ (the noise level), we first compute the median border-distance $D_{\mathrm{PSJ}}$ of PSJ after 32 iterations and the \nth{40}, \nth{50}, \nth{60} percentiles $N_{40}, N_{50}, N_{60}$ of the total amount of queries used in each attack.
We then run HSJ-$r$ for increasingly high values of $r$, which improves the median border-distance $D_{\mathrm{HSJr}}(r)$ and increases the total number of queries $N(r)$.
We stop when $D_{\mathrm{HSJr}} = D_{\mathrm{PSJ}}$ and plot the resulting ratio $N / N_{50}$ (solid line), and $N / N_{40}$, $N / N_{60}$ (small shaded area around the median line).
The result, \cref{fig:psj-hsjr}, confirms property \cref{it:p1}:
PSJ is \emph{much} more query efficient than HSJr.
At first, we were surprised that, even at very low noise levels, HSJr needed several order of magnitudes more model queries than PSJ.
The reason, we found, is that HSJ uses a very small sampling radius ($\beta=1/d$) for the gradient estimator, which impedes the estimation in the event of noise, as discussed in \cref{sec:psj-vs-hsj}. 
We therefore also compare PSJ to a version of HSJr where we replaced the original sampling radius by the same one we used in PSJ (dashed line).
The performance of HSJr improved dramatically, even though PSJ remains much more query efficient overall.
\cj{adjust here, depending on plots.}

\paragraph{Small noise breaks HSJ.}
To confirm that small noise suffices to break HSJ, we compare the performance of HSJ and PSJ on MNIST, on a deterministic classifier where labels get corrupted (flipped) with probability $\nu \in \{0,.05, .1\}$ (as in Example $\rcf_\nu$ of \cref{sec:prob-classifiers}).
Results are reported in \cref{tab:hsj-with-noise}. 
Corrupting only 1 out of 20 queries ($\nu = 5\%$) suffices to greatly deteriorate HSJ's performance (i.e., increase the median border-distance) --~even with queries repeated 3 times~--, while PSJ, with almost the same amount of queries than HSJ, is almost not affected.
Appendix~\ref{apx:hsj-with-noise} shows similar results when we replace HSJ by the boundary attack \citep{brendel18boundary}.
This inability of HSJ to deal with noise can also be seen on Figs.~\ref{fig:psj-vs-noise-model} (b) \& (d) and \ref{fig:psj-hsjr}.

%

\begin{figure}[tb]
    \centering
    \includegraphics[width=\linewidth]{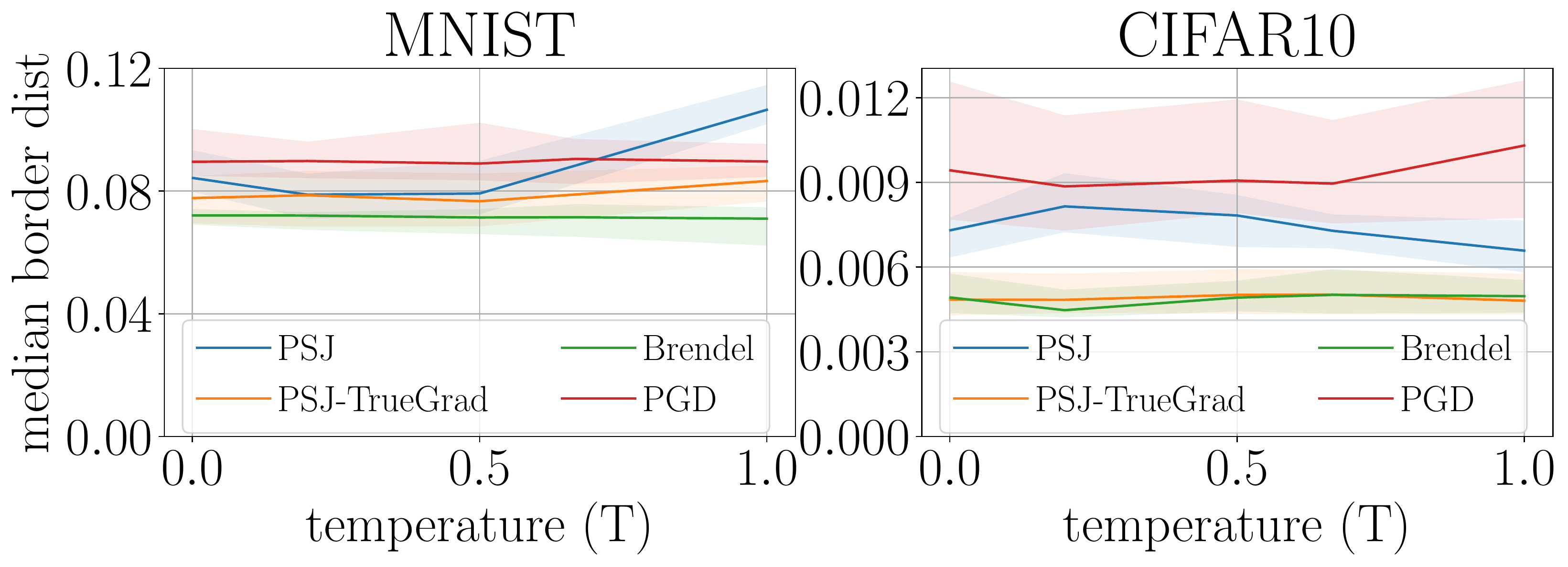}
    \caption{PSJ's adversarial examples are on par with the white-box attacks, even in noisy regimes (high $T$), and even though the white box attacks use the true gradients of the tempered logits and therefore never face any actual noise.}  
    \label{fig:whitebox}
    \vspace{-1.5em}
\end{figure}

\paragraph{PSJ vs white-box attacks.}
To evaluate how much performance we lose by ignoring information about the network architecture, we compare PSJ to several white-box attacks:
to $\ell_2$-PGD with 50 gradient steps \citep{madry18pgd} and to the $\ell_2$-attack by \citet{brendel19accurate}, using their Foolbox implementations by \citet{rauber17foolbox,rauber17foolboxnative};
and to a homemade ``PSJ-TrueGrad'' attack, where we replaced every gradient estimate in PSJ by the true gradient.
We compare these attacks on 100 MNIST and 50 CIFAR10 test images, using the ``logit sampling'' noise model at various temperatures $T$.
That way, the true underlying logits and their gradients are known and can be used by the white-box attacks without resorting to any averaging over random samples.
This trick is not applicable to other noise models and makes the comparison with PSJ doubly unfair:
first because the white-box attacks have access to the model's gradients; and second, because  here they never face any actual noise.
Given these burdens, PSJ's performance shown in \cref{fig:whitebox} is surprisingly good: it is on par with the white-box attacks.

\section{Conclusion}

\looseness -1 Although recent years have seen the development of several \emph{decision-based} attacks for \emph{deterministic} classifiers, small noise on the classification outputs typically suffices to break them.
We therefore re-designed the particularly query-efficient \emph{HopSkipJump} attack to make it work with probabilistic outputs.
By modeling and learning the local output probabilities, the resulting \emph{p}robabilistic H\emph{opSkipJump} algorithm, \emph{PopSkipJump}, optimally adapts its queries to match HSJ's performance at every iteration over increasing noise levels.
It is \emph{much} more query-efficient than the off-the-shelf alternative ``HSJ (or any other SOTA decision-based attack) with repeated queries and majority voting'', and 
matches HSJ's query-efficiency on deterministic and near-deterministic classifiers.
We successfully applied PSJ to randomized adversarial defenses proposed at major recent conferences, and showed that they offer almost no extra robustness against decision-based attack as compared to their underlying deterministic base model.
Our adaptations and statistical analysis of HSJ could be straightforwardly used to extend another decision-based attack, qFool by \citet{liu19qfool}, to cope with probabilistic answers.
Overall, we hope that our method will help crafting adversarial examples in more real-world settings with intrinsic noise, such as for sets of classifiers or for humans. 
However, our results also suggest that the feasibility of such attacks will greatly depend on the noise level, since PSJ can require orders of magnitude more queries to achieve the same performance in the probabilistic setting than in the deterministic one.

\section*{Acknowledgements}

\looseness -1 We thank the reviewers for their valuable feedback and our families for their constant support. This project is supported by the Swiss National Science Foundation under NCCR Automation, grant agreement 51NF40 180545.
CJSG is funded in part by ETH's Foundations of Data Science (ETH-FDS).



\bibliography{main}
\bibliographystyle{icml2021}

\appendix
%
%
%


\newpage
\onecolumn
\icmltitle{PopSkipJump: Decision-Based Attack for Probabilistic Classifiers\\
Supplementary Materials}

\section{Justifications and Proofs for Section~\ref{sec:grad-est}}

\subsubsection{Justifying \texorpdfstring{\cref{eq:cos}}{the cosine equation}\label{sec:justification-cos-eq}}

First, let us extend \cref{eq:sigmoid} and assume that, near the point $\vx_t$ (where we estimate the gradient), the classification probabilities $p_c(\vx)$ are given (approximately) by a planar sigmoid, meaning:
\begin{equation}\label{eq:planar-sigmoid}
    p_c(\vx) = \eps + (1-2\eps) \sigmoid(s \ipd{\vx - \vz}{\vg}) = \eps + (1-2\eps) \sigmoid(s(x - z))
\end{equation}
where $s \in (0, +\infty]$, $\vg = \vg(\vz)$ is a unit vector in the gradient direction at point $\vz$ (given by $z$ in \cref{eq:sigmoid}), and where $x, z$ are the first coordinates of $\vx, \vz$ in an orthonormal basis $\B = (\ve_1, \ve_2, ..., \ve_d)$ where $\ve_1=\vg$.
In the deterministic case, when $s=\infty$, \cref{eq:planar-sigmoid} amounts to assuming that the boundary is a linear hyperplane $\sH$ going through $\vz$ and orthogonal to $\vg$.
Also, notice that \cref{eq:planar-sigmoid} is independent of the choice of $\vz$, as long as $\vz$ is contained in the hyperplane $\sH$.

\begin{remark}[Link between \cref{eq:sigmoid} and \cref{eq:planar-sigmoid}]
\Cref{eq:planar-sigmoid} is consistent with \cref{eq:sigmoid} in the sense that $p_c(\vx)$ will indeed be a sigmoid along any arbitrary line, as in \cref{eq:sigmoid}.
However, the notations are different:
in \cref{eq:planar-sigmoid}, $s, x, z$ are coordinates along $\vg$, whereas in \cref{eq:sigmoid}, they are coordinates along the line $[\tilde{\vx}_t, \vx_*]$.
There is a factor $\cos(\tilde{\vx}_t - \vx_*, \vg)$ between the two, which, upon convergence, should converge to 1 \citep[][Thm.1]{chen19hsja}. 
\end{remark}

\begin{lemma}\label{lem:as-conv}
Let $n$ be a positive integer and $\rvdelta^{(i)} \sim \N(0, \mI_d)$ for $i=1, ... n$.
Let $\beta > 0$ and $\rcf(\vx+\beta \rvdelta^{(i)}) \sim \mathrm{Ber}(p_c(\vx+\beta \rvdelta))$ with values in $\{-1,1\}$ and where $p_c$ given by \cref{eq:planar-sigmoid}.
Let $\hat{\vg}(\vx) := \rvu / \norm{\rvu}$ with $\rvu := \sum_{i=1}^n \rcf(\vx + \beta \rvdelta^{(i)}) \rvdelta^{(i)}$.
Then
\begin{equation}\label{eq:random-cos}
    \cos(\hat{\rvg}(\vx), \vg)
    = 
    \frac{\xi}{\sqrt{\xi^2 + \chi^2_{d-1} / n}}
    \quad \text{where} \quad
    \left \{
    \begin{array}{l}
        \chi^2_{d-1} \sim \text{chi-square distribution of order } d-1\\
        \xi := \frac{1}{n} \sum_{i=1}^n \ry_\eps(s(\Delta + \beta \rdelta^{(i)})) \rvdelta^{(i)}\\
        \ry_\eps(x) \sim \mathrm{Ber}(\eps + (1-2\eps)\sigmoid(x)) \text{ and } \Delta := x-z.
    \end{array}
    \right .
\end{equation}
Moreover, when $(d, n) \to (\infty, \infty)$ with $\frac{d-1}{n}$ converging to a fixed ratio denoted $\frac{\tilde{d} - 1}{\tilde{n}}$, then, almost surely,
\begin{equation}\label{eq:cos-as}
    \cos(\hat{\rvg}(\vx), \vg)
    \overset{\mathrm{a.s.}}{\longrightarrow}
    \frac{1}{\sqrt{1 + \frac{\tilde{d}-1}{\tilde{n} \alpha}}}
    \quad \text{where} \quad
    \alpha := \E_{\rdelta, \ry}[\ry(s(\Delta + \beta \rdelta)) \rdelta] \ .
\end{equation}
\end{lemma}

\begin{figure}[tb]
    \centering
    \includegraphics[width=\linewidth]{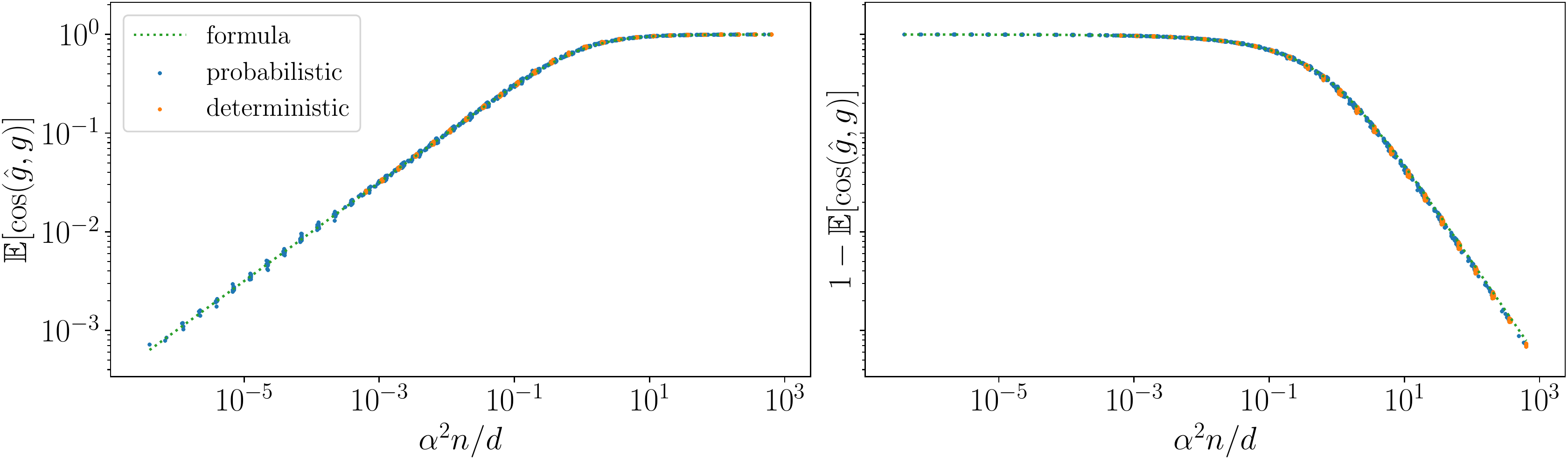}
    \caption{
        Comparing analytical and numerical approximations of $\E[\cos(\hat{\rvg}, \vg)]$.
        Dotted line: $= (1 + \frac{d-1}{n \alpha^2})^{-\nf{1}{2}}$ (eq.\ \ref{eq:cos}).
        Points: $\E[\cos(\hat{\rvg}, \vg)]$ computed by averaging repeated draws from \cref{eq:random-cos}.
        Each point represents one such average for a given combination of $(n, d, s) \in \sN \times \sD \times \sS \cup \{\infty\}$, where $\sN = \sD$ and $\sS$ where computed using numpy's $\mathtt{logspace(1,4,num=13)}$ and $\mathtt{logspace(-2, 2, num=17)}$ respectively. 
        For all points, we fixed $\beta=1$ and $x-s=0$.
        \cj{iterate over $x-s$.}
        Points for $s=\infty$ are colored orange.}
    \label{fig:cos-approx}
\end{figure}

\Cref{eq:cos-as} says that, if $n$ and $d$ are large enough, we can replace the random quantities $\xi$ and $\chi^2_{d-1}$ of \cref{eq:random-cos} by their expectations, $d-1$ and $\alpha$ respectively, and get the RHS of \cref{eq:cos}.
Hence, it proves \cref{eq:cos} in the large $n$ and $d$ limit.
However, one may wonder whether \cref{eq:cos,eq:cos-as} also hold with good approximations for finite $n,d$.
It is not difficult to estimate the order of magnitude of $n$ and $d$ needed in \cref{eq:cos-as} by computing the variances of $\xi$ and $\chi^2_{d-1}$ and using the central limit theorem.
However, since \cref{eq:cos} has an additionnal expectation on the LHS which may quicken the convergence, we will prefer to approximate $\E[\cos(\hat{\rvg}, \vg)]$ by the average over many random draws from \cref{eq:random-cos} (Monte-Carlo method) and comparing it with the results obtained by the RHS of \cref{eq:cos}.
Results are shown in \cref{fig:cos-approx}.

\begin{proof}{(\Cref{lem:as-conv}.)}
Let us work in an orthonormal basis $\B := (\ve_1 = \vg, \ve_2, ..., \ve_d)$ of $\R^d$ and possibly drop the index `$1$' for the first coordinate (as in $x$ for $x_1$).
Since $\cos(\hat{\rvg}(\vx), \vg)$ is invariant by orthogonal translations to $\vg$, i.e., by changes in the coordinates $(x_2, x_3, \ldots, x_d)$ of $\vx$, let us assume, w.l.o.g., that $\vx = (x, 0, 0, \ldots, 0)$.
Then
\begin{align*}
    \cos(\hat{\rvg}(\vx), \vg)
    &= \frac{\ipd{\hat{\vg}(\vx)}{\vg}}{\norm{\hat{\vg}} \norm{\vg}}
    \overset{(*)}{=} 
        \frac{
        \sum_{i=1}^n \rcf(\vx+\beta\rvdelta^{(i)}) \ipd{\rvdelta^{(i)}}{\vg}}{
        \norm{\sum_{i=1}^n \rcf(\vx+\beta\rvdelta^{(i)}) \rvdelta^{(i)}}} \\
    &= \frac{
        \sum_{i=1}^n \rcf(\vx+\beta\rvdelta^{(i)}) \rdelta^{(i)}}{
        \sqrt{
            \Big(
            \sum_{i=1}^n \rcf(\vx+\beta\rvdelta^{(i)}) \rdelta^{(i)} \Big)^2
        + \sum_{j=2}^d
            \Big(
            \sum_{i=1}^n
            \rcf(\vx+\beta\rvdelta^{(i)}) \rdelta^{(i)}_j
            \Big)^2}} \\
    &= \frac{\xi_1}{\sqrt{\xi_1^2 + \sum_{j=1}^d \xi_j^2}} \ ,
\end{align*}
where we defined $\xi_j := \frac{1}{n} \sum_{i=1}^n \rcf(\vx+\beta\rvdelta^{(i)}) \rdelta^{(i)}_j$.
With the change of variable $\ry(s(\Delta + \beta \rdelta^{(i)})) = \rcf(\vx + \beta \rvdelta^{(i)})$ and using \cref{eq:planar-sigmoid}, we see that $\ry(x) \sim \eps + (1-\eps) \sigmoid(x)$ and get $\xi_j := \frac{1}{n} \sum_{i=1}^n \ry(s(\Delta + \beta \rdelta^{(i)})) \rvdelta^{(i)}$.

Since, for any $i$, $\ry(s(\Delta + \beta\rdelta^{(i)}))$ follows a Bernoulli distribution that is independent of $\rdelta^{(i)}_j \sim \N(0, 1)$ whenever $j \geq 2$, the products $\ry(s(\Delta + \beta\rdelta^{(i)})) \rdelta^{(i)}_j$ follows again a standard normal distribution.
So, for any $j \geq 2$, $\xi_j$ is the mean of $n$ independent Gaussians $\N(0, 1)$, hence $\xi_j \sim \N(0, \sigma^2 = 1 / n)$.
Since all $(\rdelta^{(i)}_j)_{ij}$ are mutually independent, so are the products $(\ry(s(\Delta + \beta\rdelta^{(i)})) \rdelta^{(i)}_j)_{ij}$, and therefore also all $(\xi_j)_j$.
Hence $\chi^2_{d-1}  := n \sum_{j=2}^d \xi_j^2$ follows a chi-squared distribution of order $d-1$, which yields \cref{eq:random-cos}, where $\xi = \xi_1$.

For \cref{eq:cos-as}, note that, by the law of large numbers, almost surely, $\xi_1 \to \E[\xi_1] = \alpha$ as $n \to \infty$, and $\nf{\chi^2_{d-1}}{d-1} \to 1$ as $d \to \infty$, i.e., $\nf{\chi^2_{d-1}}{n} = \frac{d-1}{n}\frac{\chi^2_{d-1}}{(d-1)} \to \frac{\tilde{d}-1}{\tilde{n}}$.
We conclude by applying the continuous mapping theorem with the function $(x_1,x_2) \mapsto x_1/\sqrt{x_1^2 + x_2}$ to $(\xi_1, \nf{\chi^2_{d-1}}{n})$ when $n, d \to \infty$ with $\frac{d-1}{n} \to \frac{\tilde{d}-1}{\tilde{n}}$.
\end{proof}

\subsubsection{Proof of \texorpdfstring{\cref{prop:alpha}}{alpha equation}\label{sec:proof-alpha-eq}}

First, the following computations shows that we can recover the case of arbitrary $\beta>0$ from the case $\beta=1$.
\begin{equation*}
    \alpha(\Delta, s, \beta, \eps)
    := \E[\ry_\eps(s(\Delta+\beta \rdelta)) \rdelta]
    = \E[\ry_\eps(s\beta (\Delta/\beta+\rdelta)) \rdelta]
    = \alpha(\Delta/\beta, s\beta, 1, \eps)
\end{equation*}
So, from now on, let us assume that $\beta = 1$.
Then
\begin{align*}
    \E[\ry_\eps(s(x+\rdelta))\rdelta]
    &= \int_{\delta=-\infty}^{+\infty} \delta
        \Big(
        \underbrace{
        p(\rdelta = \delta, \, \ry_\eps(s(\Delta + \rdelta))=1)}_{
        p(\rdelta = \delta) (\eps + (1-2\eps)\sigmoid(s(\delta + \Delta)))}
        +
        \underbrace{
        p(\rdelta = -\delta, \, \ry_\eps(s(\Delta + \rdelta))=-1)}_{
        p(\rdelta = \delta) (1 - \eps - (1-2\eps)\sigmoid(s (-\delta + \Delta)))}
        \Big)
        \diff \delta \\
    &= \int_{\delta=-\infty}^{+\infty}
        \delta \N(\delta; 0 ,1)
        \Big(
        (1-2\eps)\sigmoid(s (\delta + \Delta) ) + 1 - (1-2\eps)\sigmoid(s (\Delta - \delta))
        \Big)
        \diff \delta \\
    &= (1-2\eps) \Bigg(
        \underbrace{
        \int_{\delta=-\infty}^{+\infty} \delta \N(\delta; 0 ,1)
        \sigmoid(s (\delta + \Delta) ) \diff \delta}_{(**)}
        + 
       \underbrace{
       \int_{\delta=-\infty}^{+\infty} \delta \N(\delta; 0 ,1)
        \sigmoid(s (\delta - \Delta) ) \diff \delta}_{(*)}
        \Bigg) \ .
\end{align*}
Next we compute $(*)$ and $(**)$.
\begin{align*}
    (*)
    &= \int_{-\infty}^{\Delta-\nf{1}{2s}} 0 \diff \delta
    + \underbrace{
      \int_{\Delta-\nf{1}{2s}}^{\Delta+\nf{1}{2s}}
        \delta \N(\delta; 0, 1) (\nf{1}{2} + s(\delta - \Delta)) \diff \delta}_{(a)}
    + \underbrace{
      \int_{\Delta+\nf{1}{2s}}^{+\infty} \delta \N(\delta; 0, 1) \diff \delta}_{(b)}\\
    (a)
    &= 
    \frac{1}{\sqrt{2 \pi}}
    \int_{\Delta - \nf{1}{2s}}^{\Delta + \nf{1}{2s}}
    s (\delta^2-1) e^{\delta^2/2}
    + s e^{\delta^2/2} 
    + (\nf{1}{2} - s \Delta) \delta e^{-\Delta^2/2} \, \diff \delta \\
    &=
    \frac{1}{\sqrt{2 \pi}} 
    \left[
    -s \delta e^{-\delta^2/2}
    + s \sqrt{2\pi} \Phi(\delta)
    - (\nf{1}{2} - s \Delta) e^{-\delta^2/2}
    \right]_{\Delta-\nf{1}{2s}}^{\Delta+\nf{1}{2s}} \\
    (b) 
    &=
    \frac{1}{\sqrt{2 \pi}} e^{-\frac{1}{2}(\frac{1}{2s} + \Delta)^2} \ ,
\end{align*}
where $\Phi$ designates the cumulative distribution function of the standard normal.
Noticing that $(**)$ is $(*)$ with $\Delta$ replaced by $-\Delta$ (in $(a)$ and $(b)$) and adding everything up, we get
\begin{align}
    \E[\ry_\eps(s(\Delta + \rdelta))\rdelta]
    &= (1-2\eps) 2s \left(\Phi(\nf{1}{2s} + \Delta) + \Phi(\nf{1}{2s} - \Delta) - 1 \right) \nonumber\\
    &= (1-2\eps) 2s \left(\Phi(\nf{1}{2s} + \Delta) + \Phi(\Delta - \nf{1}{2s}) \right) \nonumber\\
    &= (1-2\eps) s
    \left(
    \erf \left( \frac{\Delta + \nf{1}{2s}}{\sqrt 2} \right)
    -
    \erf \left( \frac{\Delta - \nf{1}{2s}}{\sqrt 2} \right)
    \right) \ . \label{eq:alpha}
\end{align}

As for the (deterministic) case $s=\infty$, we can either redo the previous calculations 
with $\sigmoid$ being the step function $\1(x) = 1$ if $x \geq 0$ and $0$ otherwise;
or we can let $s \to \infty$ in \cref{eq:alpha} and use a Taylor development the error-function $\erf$ centered on $\nf{\Delta}{\sqrt 2}$, which gives
\begin{align*}
    \E[\ry_\eps(s(\Delta+\rdelta))\rdelta]
    = (1-2\eps) s 
    \left(
    \erf'\left(\frac{\Delta}{\sqrt 2}\right) \frac{1}{s \sqrt{2}}
    +
    \erf'''\left(\frac{\Delta}{\sqrt 2}\right) \frac{2}{3! (2s\sqrt 2)^3}
    +
    O(\frac{1}{s^5})
    \right)
\end{align*}
with $\erf'(x) = \frac{2}{\sqrt \pi} e^{-x^2}$ and $\erf'''(x) = \frac{4}{\sqrt \pi} (2 x^2 -1) e^{-x^2}$.
Hence
\begin{align*}
    \E[\ry_\eps(s(\Delta+\rdelta))\rdelta]
    =
    (1-2\eps) \sqrt{\frac{2}{\pi}} e^{-\Delta^2/2}
    \left(
    1 + (\Delta^2-1) \frac{1}{24s^2} + O\left(\frac{1}{s^4}\right)
    \right)
    \ \rightarrow \
    (1-2\eps) \sqrt{\frac{2}{\pi}} e^{-\nf{\Delta^2}{2}} . \tag*{\qedhere}
\end{align*}

\newpage

\section{Additionnal Plots and Tables}

\subsection{Extension of \cref{tab:hsj-with-noise}}\label{apx:hsj-with-noise}

In \cref{tab:hsj-with-noise-dist,tab:hsj-with-noise-calls} we extend \cref{tab:hsj-with-noise} on the performance of decision-based attacks in presence of small noise.
The extended tables also show the performance of the boundary attack \citep{brendel18boundary} (BA) and of the boundary attack with three repeated queries and majority voting (BA-repeat3).
We have now broken the results into two tables:
\cref{tab:hsj-with-noise-dist} shows the performance of the various attacks in terms of border-distance (see definition \cref{sec:experiments});
where \cref{tab:hsj-with-noise-calls} reports the total number of model calls needed.
Overall, both tables confirm \cref{tab:hsj-with-noise}:
small noise suffices to break traditional decision-based attacks, even with repeated queries.
The performance of PSJ stays identical, with only a few more queries in the noisy setting (and far less queries than the repeated queries based attacks).

\begin{table*}[tbh]
\caption{Extension of \cref{tab:hsj-with-noise}, showing the median border-distance achieved by various decision-based attacks at various noise levels. 
PSJ, HSJ, BA stand for PopSkipJump (our attack), HopSkipJump and boundary attack \citep{brendel18boundary} respectively.
HSJ-repeat3 and BA-repeat3 are HSJ and BA where we repeat every  query 3 times and apply majority voting.
First number reports the median border-distance, as defined in \cref{sec:experiments}.
The two numbers in bracket are the \nth{40} and \nth{60} percentiles.
All numbers were computed on the MNIST test subset of 500 images using the CNN described in \cref{sec:experiments}.
A noise level $\nu=5\%$ means that the CNN outputs its argmax label with probability $.95$, and some other random label with probability $.05$.
Conclusion: all attacks perform similarly in the deterministic case ($\nu=0$), but small noise $\nu$ suffices to break HSJ and BA (with or without repeated queries), but not PSJ. }
\label{tab:hsj-with-noise-dist}
\begin{center}
\begin{tabular}{|c|c|c|c|c|c|}
\hline
\textbf{FLIP}  &\textbf{PSJ} &\textbf{HSJ}  &\textbf{HSJ-repeat3}  &\textbf{BA}  &\textbf{BA-repeat3} \\  %
\hline
$\nu=0\%$	&0.60[0.45-0.72]	&0.61[0.51-0.72]	&0.59[0.48-0.72]	&0.46[0.38-0.55]	&0.49[0.36-0.56]	\\
$\nu=5\%$	&0.63[0.53-0.76]	&2.16[1.76-2.93]	&0.75[0.58-1.01]	&15.58[13.66-16.94]	&15.60[12.94-16.73]	\\
$\nu=10\%$	&0.65[0.55-0.80]	&3.67[3.15-4.26]	&1.32[0.98-1.61]	&15.84[14.35-16.85]	&15.90[14.24-16.96]	\\
\hline
\end{tabular}
\end{center}
\end{table*}

\begin{table*}[tbh]
\caption{Same as \cref{tab:hsj-with-noise-dist} but showing the median model calls.
In the deterministic setting, PSJ only needs a few more calls than HSJ, and much less calls than BA.
In the noisy setting, the number of calls increase only slightly for PSJ, while its performance (\cref{tab:hsj-with-noise-dist}) is often much better than the repeated-query attacks, which need much more calls.}
\label{tab:hsj-with-noise-calls}
\begin{center}
\begin{tabular}{|c|c|c|c|c|c|}
\hline
\textbf{FLIP}  &\textbf{PSJ} &\textbf{HSJ}  &\textbf{HSJ-repeat3}  &\textbf{BA}  &\textbf{BA-repeat3} \\
$\nu=0\%$	&14270	&12844	&38532	&27560	&82665	\\
$\nu=5\%$	&17856	&13333	&42009	&27561	&82716	\\
$\nu=10\%$	&17931	&13108	&40813	&27551	&82677	\\
\hline
\end{tabular}
\end{center}
\end{table*}

\FloatBarrier

\newpage

\subsection{From PSJ to HSJ for various attacks noise levels\label{sec:psj-to-hsj-complete}}

\begin{figure*}[!h]
    \centering
    \includegraphics[width=\linewidth]{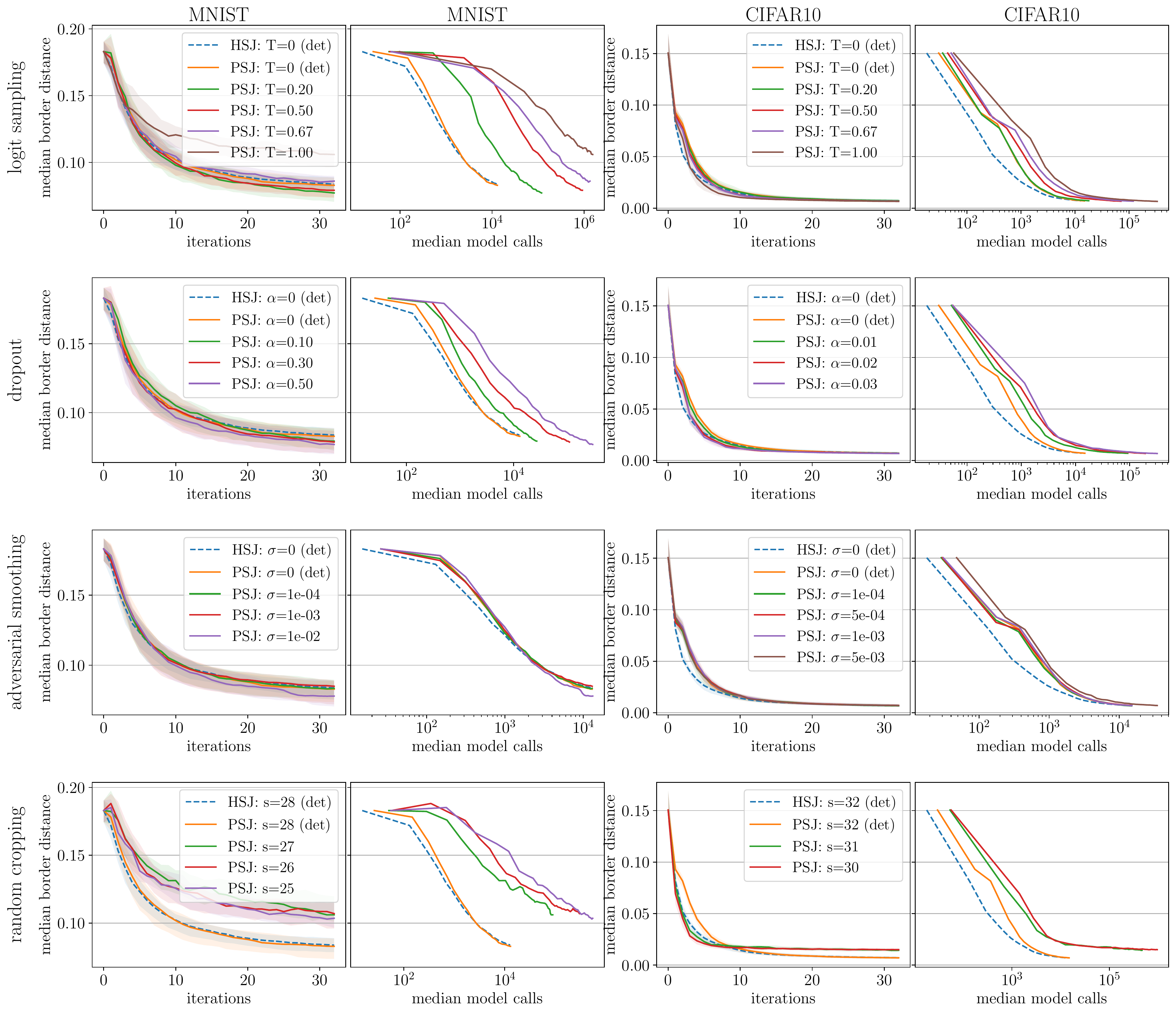}
    \caption[PSJ to HSJ]{Evolution of PopSkipJump's performance (median border-distance) with the amount of algorithm iterations (columns a \& c) and model queries (columns b \& d).  
    Each row uses a different noise model (i.e., randomization scheme) applied to a same deterministic base classifier (CNN for MNIST, Densenet for CIFAR10).
    Each curve uses a different noise level.
    Columns a \& c illustrate property \cref{it:p2}:
    the \emph{per iteration} performance of PSJ is largely independent of the noise level and noise model, and is on par with HSJ's performance on the deterministic base classifier.
    Columns b \& d illustrate property \cref{it:p1}:
    when the noise level decreases and the classifier becomes increasingly deterministic, the \emph{per query} performance of PSJ converges to that of HSJ, in the sense that the PSJ curves become more and more similar to the limiting HSJ curve.
    \label{fig:psj-vs-noise-level-complete}}
\end{figure*}

\FloatBarrier

\newpage

\subsection{Output probabilities along bin-search lines are sigmoids\label{sec:sigmoids}}

\begin{figure}
    \centering
    \includegraphics[clip, trim=1.5cm 3cm 1.5cm 3cm, width=.7\linewidth]{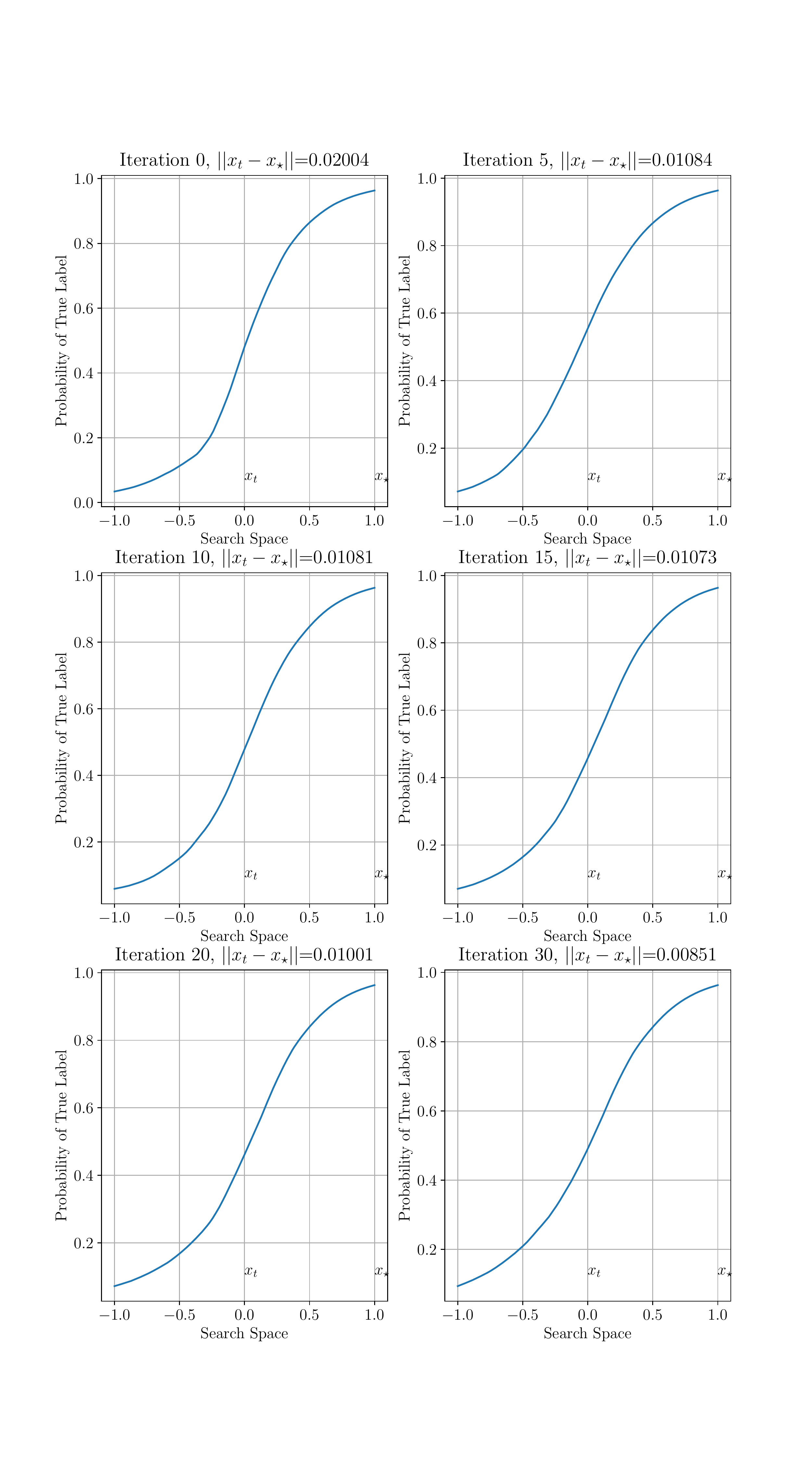}
    \caption{Output probabilities of the CNN along binary search direction $[\tilde{\vx}_t, \vx_*]$ for different iterations $t$ of an attack on some fixed MNIST image $\vx_*$.
    These plots match our assumption that the model's probabilities along binary search directions are given by a sigmoid, as in \cref{eq:sigmoid}.
    This assumption was satisfied for almost all images $\vx_*$ that we checked.}
    \label{fig:sigmoids-mnist}
\end{figure}

\begin{figure}
    \centering
    \includegraphics[clip, trim=1.5cm 3cm 1.5cm 3cm, width=.7\linewidth]{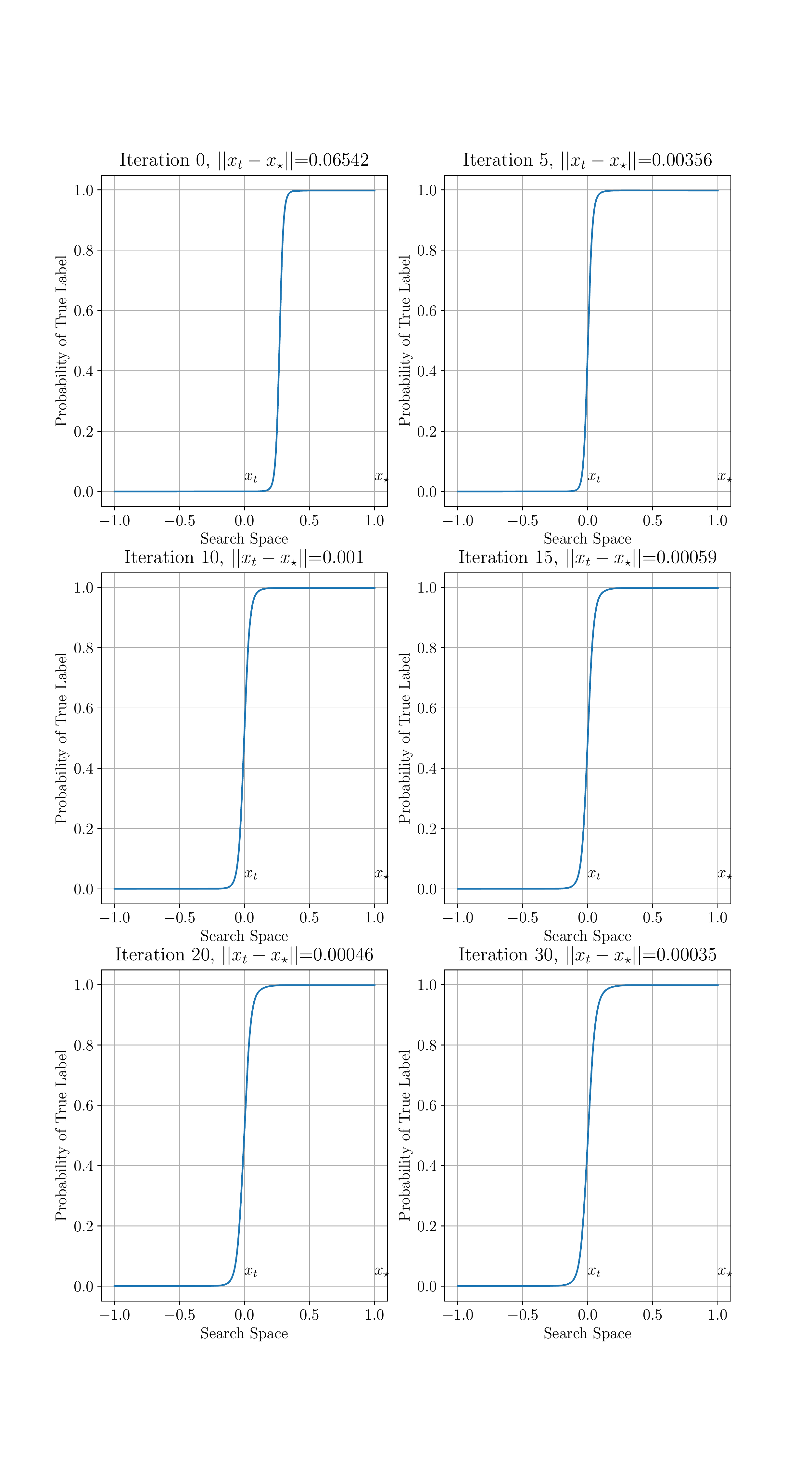}
    \caption{Same as \cref{fig:sigmoids-mnist}, but for the Densenet used on a CIFAR-10 image $\vx_*$.
    There again, the output probabilities sigmoidal, but closer to a step-function than in \cref{fig:sigmoids-mnist}.}
    \label{fig:sigmoids-cifar}
\end{figure}

In this section we briefly corroborate our assumption from \eqref{eq:sigmoid} that the probabilities of neural networks along the binary search lines $[\vx_t, \vx_*]$ have a sigmoidal shape.
We do so by plotting these probabilities in \cref{fig:sigmoids-mnist,fig:sigmoids-cifar} on two randomly chosen images --~one from MNIST and one from CIFAR10~-- at various iterations of the attack.
Note that we got similar plots for almost all images that we tested.

\FloatBarrier

\newpage

\section{How sensitive is HSJ to the number of gradient queries $n_t$ per iteration?}

\begin{figure}[htb]
    \centering
    \includegraphics[width=\linewidth]{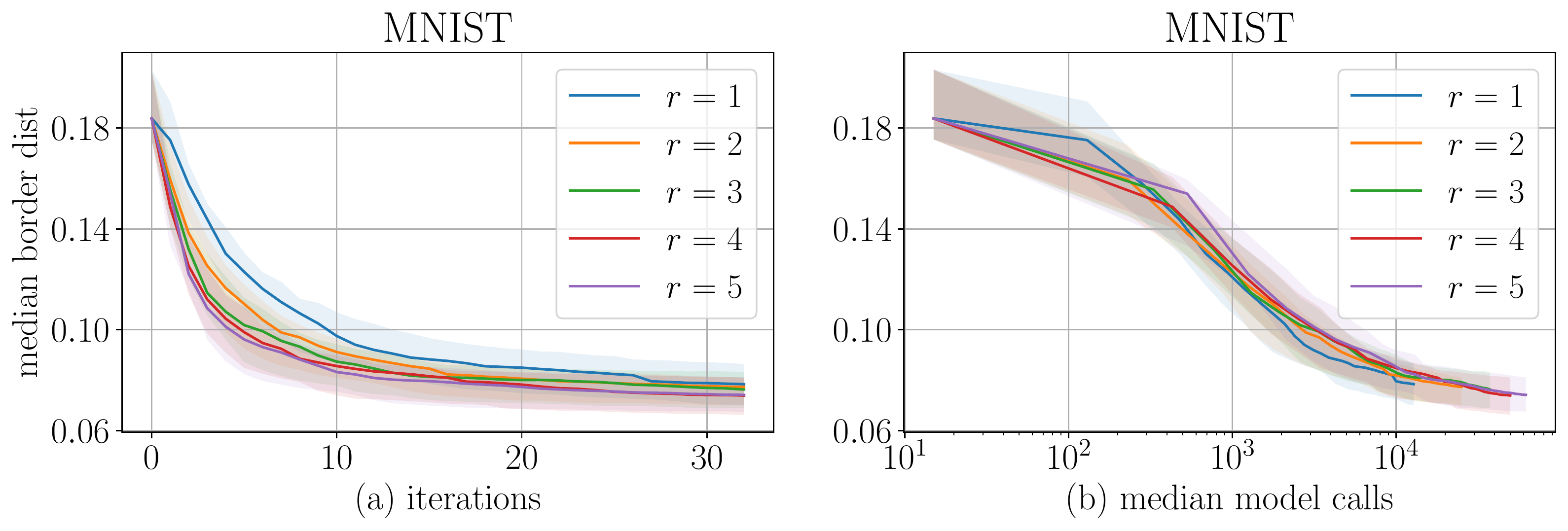}
    \caption{Performance of HSJ (lower is better) on a deterministic classifier when multiplying the default number of gradient queries $n_t$ at every iteration $t$ by a constant factor $r$. 
Not surprisingly, the performance as a function of the algorithm's iterations (Fig.\ a) increases with the query factor $r$, since more queries means a better gradient estimate.
However, the performance as function of the overall number of queries (Fig.\ b) seems independent of $r$, which suggests that the considerations in \cref{sec:grad-est} and \cref{eq:cos,eq:n}, however beautiful in theory, are not essential to the query efficiency and success of PSJ.}
    \label{fig:hsj-increased-sample-size}%
\end{figure}

In \cref{sec:grad-est} we showed how to use the estimated output distribution from the binary search step to compute the number of gradient queries required to get the same estimation quality in the noisy case than in the deterministic one.
This makes PSJ's per-iteration performance independent of the noise level 
and would in particular allow us to optimize the number of gradient queries in the deterministic setting only, and then automatically infer the optimal number of queries for all noisy settings without any additional optimization.
One natural question then, however, is how sensitive HSJ is to the number of gradient queries $n_t$ per round $t$.
To get a rough idea, we plot the performance of HSJ in the deterministic setting when multiplying the original, default number of gradient queries $n_t = 100 \sqrt t$ by a factor $r=1, 2, ..., 5$.
The results are shown in \cref{fig:hsj-increased-sample-size}.

Not surprisingly, the performance \emph{per iteration} of HSJ (Fig.\ a) increases with $r$, since more queries means a better gradient estimate.
However, the performance as a function of the overall number of queries (Fig.\ b) seems independent of $r$.
This suggests that the overall performance of HSJ --~and therefore also of PSJ~-- is not very sensitive to the actual number of gradient queries per iteration, which in turn suggests that the considerations in \cref{sec:grad-est} and formulae \cref{eq:cos,eq:n} are not essential to the query efficiency and success of PSJ.

\newpage

\section{Time, query and computational complexity of PopSkipJump\label{sec:complexity}}

\begin{figure}[htb]
    \centering
    \includegraphics[width=\linewidth]{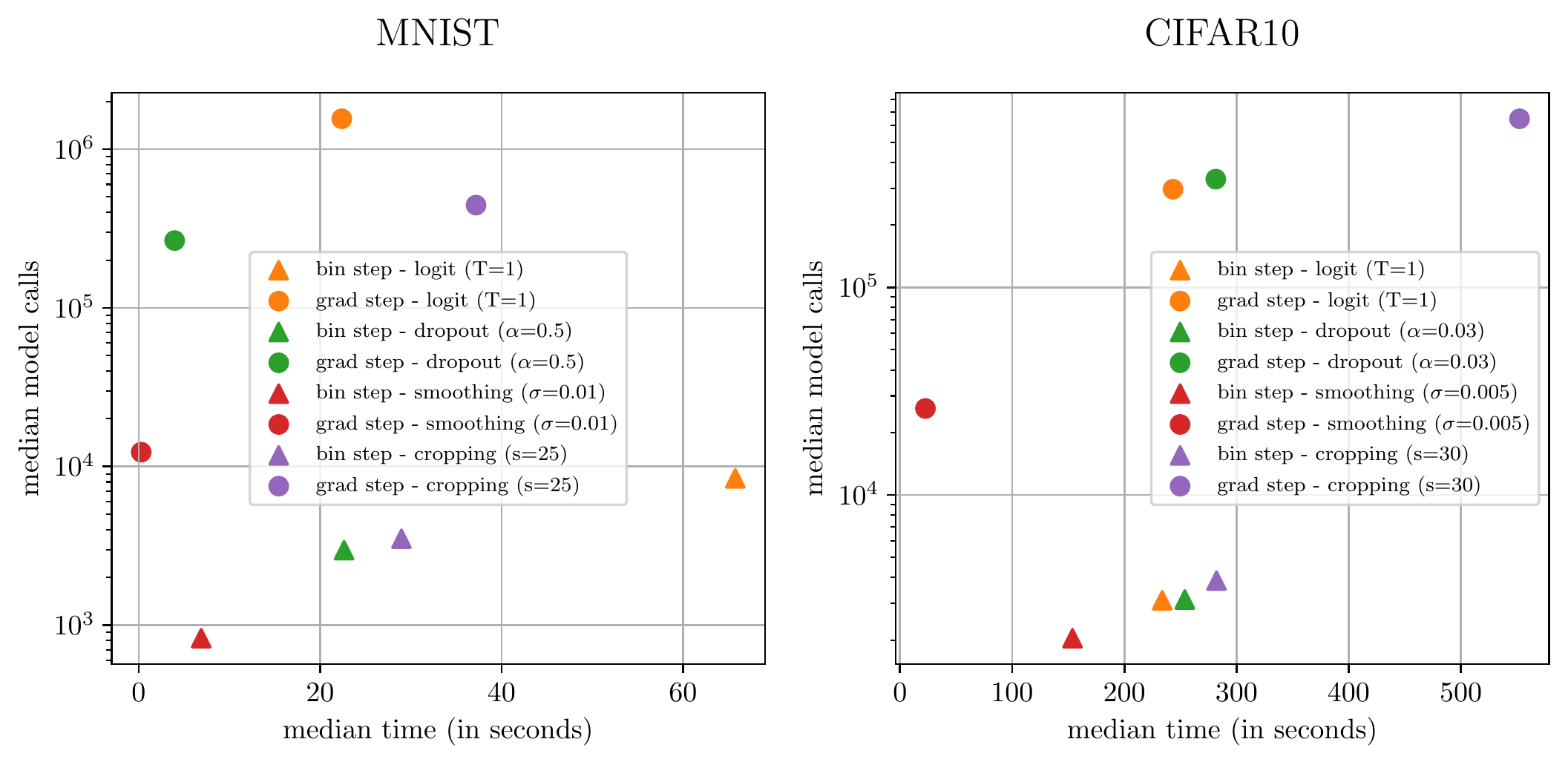}
    \caption{Number of queries versus time spent in different parts of the PSJ algorithm.
    Gradient estimation needs many ($\geq 100\times$) more queries than noisy bin-search but less time, because its queries are independent, hence batchable and parallelizable, whereas noisy bin-search is sequential and has an expensive information maximization step after each query.
    Here, gradient estimation queries are computed in batches of 256 queries (on a single GPU).
    For every noise model, we used the highest noise levels that we considered in this paper (temperature $T$, dropout rate $\alpha$, standard deviation $\sigma$ and cropped size $s$).%
    \label{fig:queries-vs-time}%
    }
\end{figure}

\paragraph{Overview.}
PopSkipJump has two resource-intensive parts: noisy binary search and the gradient estimation step (points \ref{it:bin-search} and \ref{it:grad-est} in \cref{sec:HSJ} respectively).
Gradient estimation typically needs many more queries than binary search ($\geq 100\times$ more).
Every query has same time and computational complexity, which increases with the network architecture.
But the queries for gradient estimation are all independent and can therefore be batched and parallelized.
The overall time for gradient estimation can in principle be driven down arbitrarily with enough cores or GPUs.
Noisy bin-search on the other side is sequential by essence and has an expensive mutual information maximization step after every query.
In our experiments, this leads to comparable time complexities, as shown in \cref{fig:queries-vs-time}.
We will now first discuss in more details the \emph{query} complexities of bin-search and gradient estimation, then the \emph{computational} complexity of the information maximization step, and finish with two tricks to accelerate the noisy bin-search steps.

\subsection{Query complexities.}

Both for the noisy bin-search and for the gradient estimation, the number of queries depends mainly on the true parameters $\eps$ and $s$ of the underlying sigmoid (i.e., roughly speaking, on the noisiness of the classifier):
increasing the noise $\eps$ and/or decreasing the shape $s$ (i.e., flattening the sigmoid) tends to increase the expected amount of queries.
For gradient estimation, the exact number of queries is computed as described by \cref{alg:psj} and its formula for $n_t$ (see also \cref{sec:grad-est}).
There, for the computation of $C^{det}_t$ (the expected cosine at step $t$ for a deterministic classifier), we used $n^{det}_t = 100 \sqrt{t}$ (as in the original HSJ algorithm), and $\theta^{det}_t = 0.010$ and $\beta = 0.280$ for MNIST, and $\theta^{det}_t = 0.003$ and $\beta = 0.185$ for CIFAR10.
(As explained in the caption of \cref{fig:various-deltas}, since we use a larger $\beta$ than in the original HSJ algorithm, we also use a larger ``bin-size'' $\theta^{det}_t$.)
As for noisy bin-search, the number of queries is driven by the number of queries needed to determine the posterior distribution of the sigmoid parameters with sufficient precision to meet the stopping criterium from \cref{sec:grad-est}.
More noise $\eps$ and flatter sigmoid shapes (i.e., smaller values of $s$) both decrease the expected amount of information on the sigmoid's center $z$ carried by every query, which increases the expected total number of queries needed.

\subsection{Computational complexity.}

As discussed above, the computational complexity of every query --~be it for bin-search or for gradient estimation~-- is mainly driven by the network architecture and its size.
The complexity of the mutual information maximization, however, depends on the discretizations used to model the prior/posterior probabilities of the sigmoid parameters $z$, $\eps$ and $s$.
More precisely, we represent our priors/posteriors over $z$, $\eps$ and $s$ by constraining them to intervals $I_z$, $I_\eps$, $I_s$ that are discretized into $n_z$, $n_\eps$ and $n_s$ bins/points respectively.
At every bin-search step, and for $n_x$ values of $x \in I_z$, we compute the mutual information $\info{\rcf(x)}{\rs,\rz,\reps}$ given by:
\[
    \info{\rcf(x)}{\rs,\rz,\reps} =
        \sum_{z \in I_z, \eps \in I_\eps, s \in I_s} p(\rcf(x), s, z, \eps) \log \frac{p(\rcf(x), s, z, \eps)}{p(\rcf(x)) p(s, z, \eps)}
\]
where $p(\rcf(x), s, z, \eps) =  p(\rcf(x) \, | \, s, z, \eps) p(s, z, \eps) = p_c(x) p(s, z, \eps)$ (see \cref{eq:sigmoid}), $p(\rcf(x))$ is given by marginalizing out $s, z, \eps$ in $p(\rcf(x), s, z, \eps)$ and where $p(s, z, \eps)$ is the current prior/posterior.
Hence, $\info{\rcf(x)}{\rs,\rz,\eps}$ is a sum of $n_z n_\eps n_s$ terms with same complexity each, so it costs $O(n_z n_\eps n_s)$.
Since we repeat this computation for every location $x \in I_z$, the overall computation of the mutual information acquisition function has complexity
\begin{equation}\label{eq:complexity-mutual-info}
    \text{complexity of mutual info maximization} = O(n_x n_z n_\eps n_s).
\end{equation}
In practice, we chose $I_\eps = \{0.9, 1\}$ (i.e., $n_\eps = 2$), and $\log_{10}(I_s) = [1,3]$ with $n_s = 31$, and
\begin{itemize}[nosep]
    \item for MNIST, $I_z = I_x = [0, 1]$ with $n_z = n_x = 101$;
    \item for CIFAR10, $I_z = I_x = [0, 1]$ with $n_z = n_x = 301$.
\end{itemize}
The values of $n_z$ were chosen so that the step-size $\tilde{\theta}$ would roughly respect the ratio $\beta / \tilde{\theta} = \delta_t / \theta_t = \sqrt{d}$ suggested by \citet{chen19hsja} \ns{reference not working?}.
They could probably be optimized further.

\subsection{Accelerating noisy binary search}

In this section we study the following two tricks to accelerate the noisy bin-search steps. 
\begin{enumerate}[label=(\arabic*)]
\item \emph{sampling multiple points after each information maximization step}:\label{it:acc1}
since one of the bottle-necks of noisy bin-search is that queries cannot be batched, we considered altering the algorithm by querying the classifier multiple times after every mutual information maximization step.
These multiple queries could occur either at the same point $x$ --~the maximizer of the mutual information~--, or uniformly over all points that are within a range of, say, $90\%$ of the maximum.
Thereby, one may lose some query efficiency (more queries needed for a same average information gain), but spare a lot of time via query batching.
Since when writing this paper, we were primarily thinking of applications were query efficiency mattered most, we did not include this acceleration trick in our experiments.
However, in many other applications, using a bit more queries to save wall-clock time of computation can be the better option.

\item \emph{reducing the range of priors}:\label{it:acc2}
we noticed that the parameters of the sigmoid found by the noisy bin-search procedure become increasingly similar from one PSJ iteration to other.
So, instead of re-initializing the priors uniformly over the same intervals $I_z$ and $\log_{10}(I_s)$ at the beginning of every bin-search procedure, we used intervals $\tilde{I}_z$ and $\log_{10}(\tilde{I}_s)$ that were centered on the output $\hat{z}$ and $\log_{10}{\hat{s}}$ of the previous iteration and whose length we decreased at every iteration.
We chose this length to be a fraction $1/k$ of the length of the original intervals $I_z$ and $\log_{10}(I_s)$, with $k=1\ldots10$ for iterations 1 to 10, and $k=10$ for iterations $\geq 10$.
\end{enumerate}

\begin{figure}[htb]
    \centering
    \includegraphics[width=\linewidth]{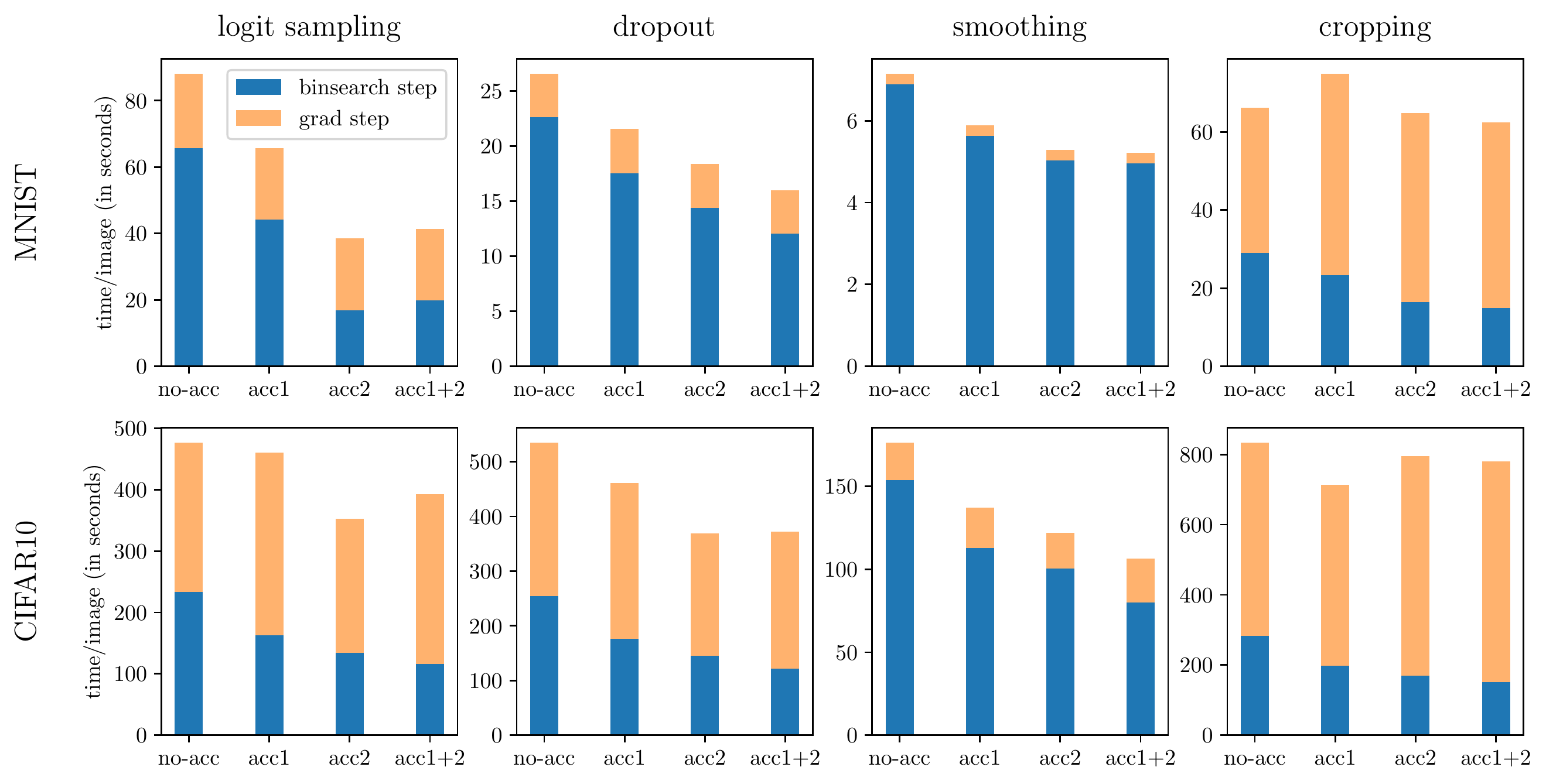}
    \caption{Median time (over 20 images) spent for bin-search and gradient estimation with and without acceleration tricks.
    As expected, the acceleration tricks \ref{it:acc1} and \ref{it:acc2} do help accelerating the binary search step.  
    Note that the acceleration tricks should not affect the gradient estimation step.
    For the multiple sampling trick \ref{it:acc1}, we made 5 queries per information maximization step.}
    \label{fig:timing_acceleration}%
\end{figure}

\begin{figure}[htb]
    \centering
    \includegraphics[width=\linewidth]{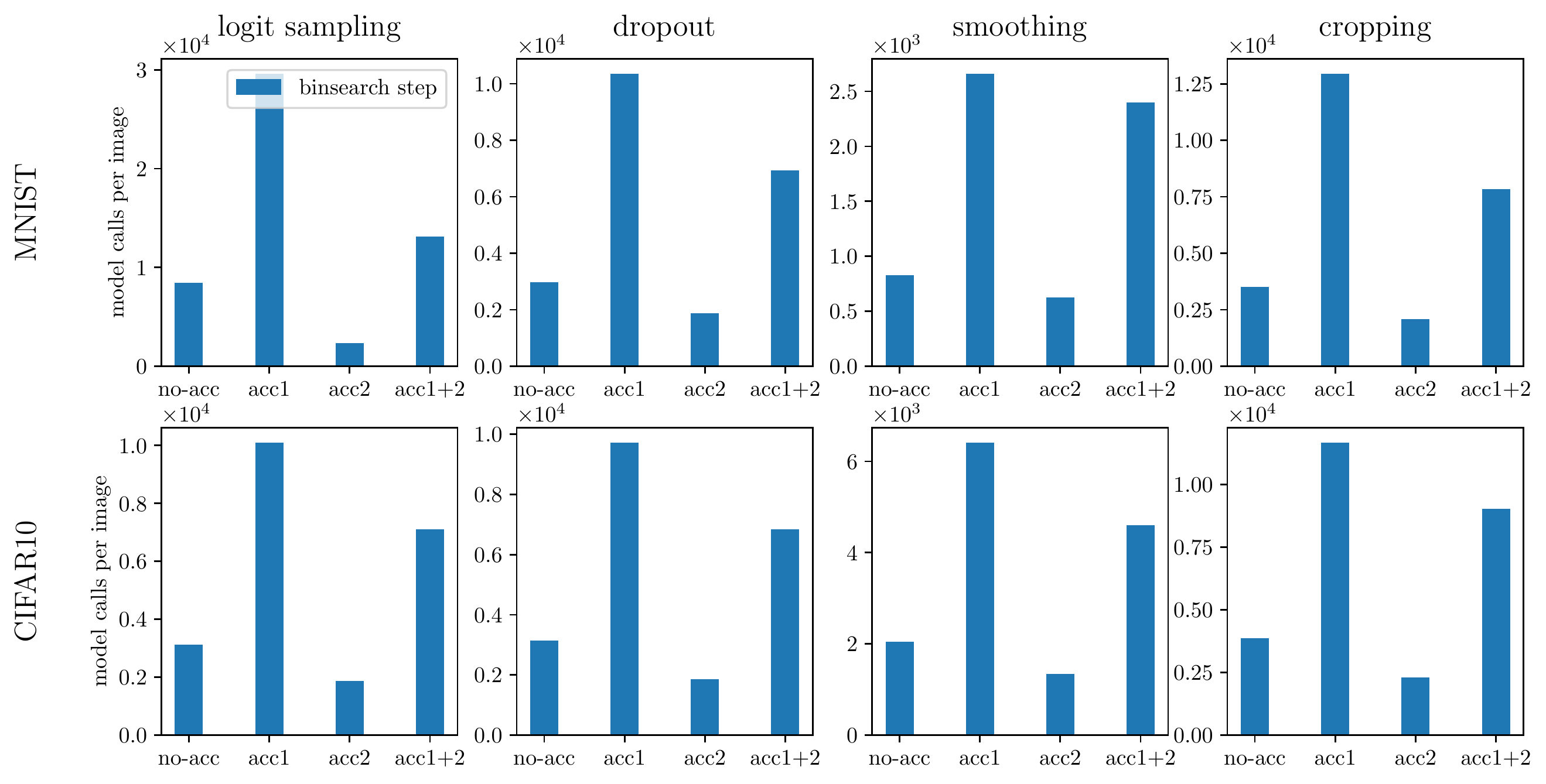}
    \caption{Median amount (over 20 images) of total bin-search queries with and without acceleration tricks.
    Acceleration trick \ref{it:acc1} (multiple queries after each information maximization) increases the number of queries, since the expected amount of information conveyed by each query is reduced.
    Acceleration trick \ref{it:acc2} (tightening the priors on the sigmoid's parameters) reduce the amount of bin-search queries.
    Note however that the number of noisy bin-search queries is 2 orders of magnitudes smaller than the number of queries for gradient estimations.
    So the variations observed here have almost affect the overall amount of queries per attack.
    For the multiple sampling trick \ref{it:acc1}, we made 5 queries per information maximization step.}
    \label{fig:queries_acceleration}%
\end{figure}

\begin{figure}[htb]
    \centering
    \includegraphics[width=\linewidth]{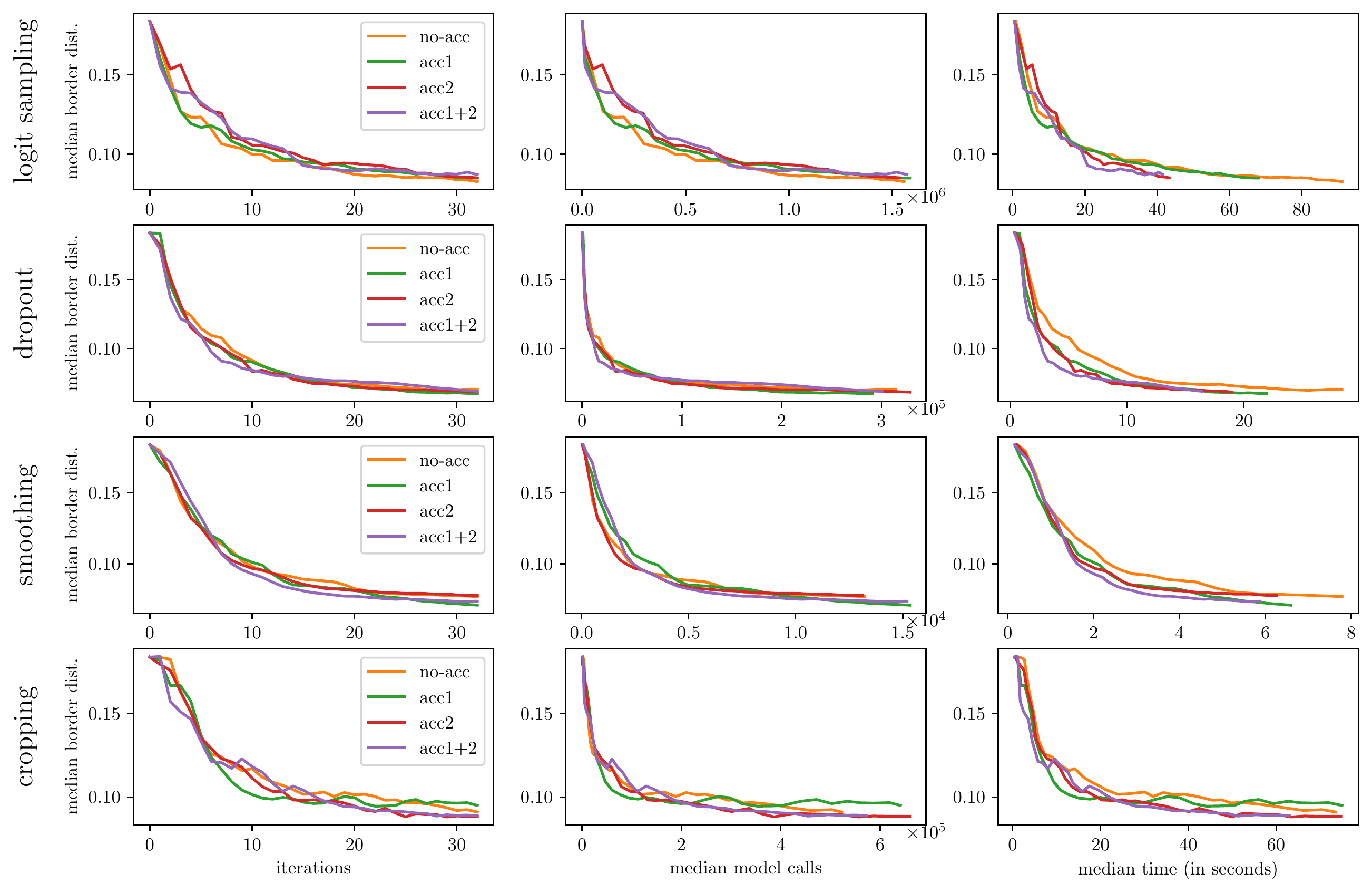}
    \caption{\textbf{MNIST}: Effects of acceleration tricks.
    Column (a) shows that acceleration has no significant effect on the algorithm's output (i.e., on the median border distance after every iteration).
    Column (b) shows that, as expected, the algorithm runs faster (in wall-clock time), with a similar output performance.
    Column (c) shows that the median number of model calls is not significantly affected by the acceleration.}
    \label{fig:acceleration_mnist}%
\end{figure}

\begin{figure}[htb]
    \centering
    \includegraphics[width=\linewidth]{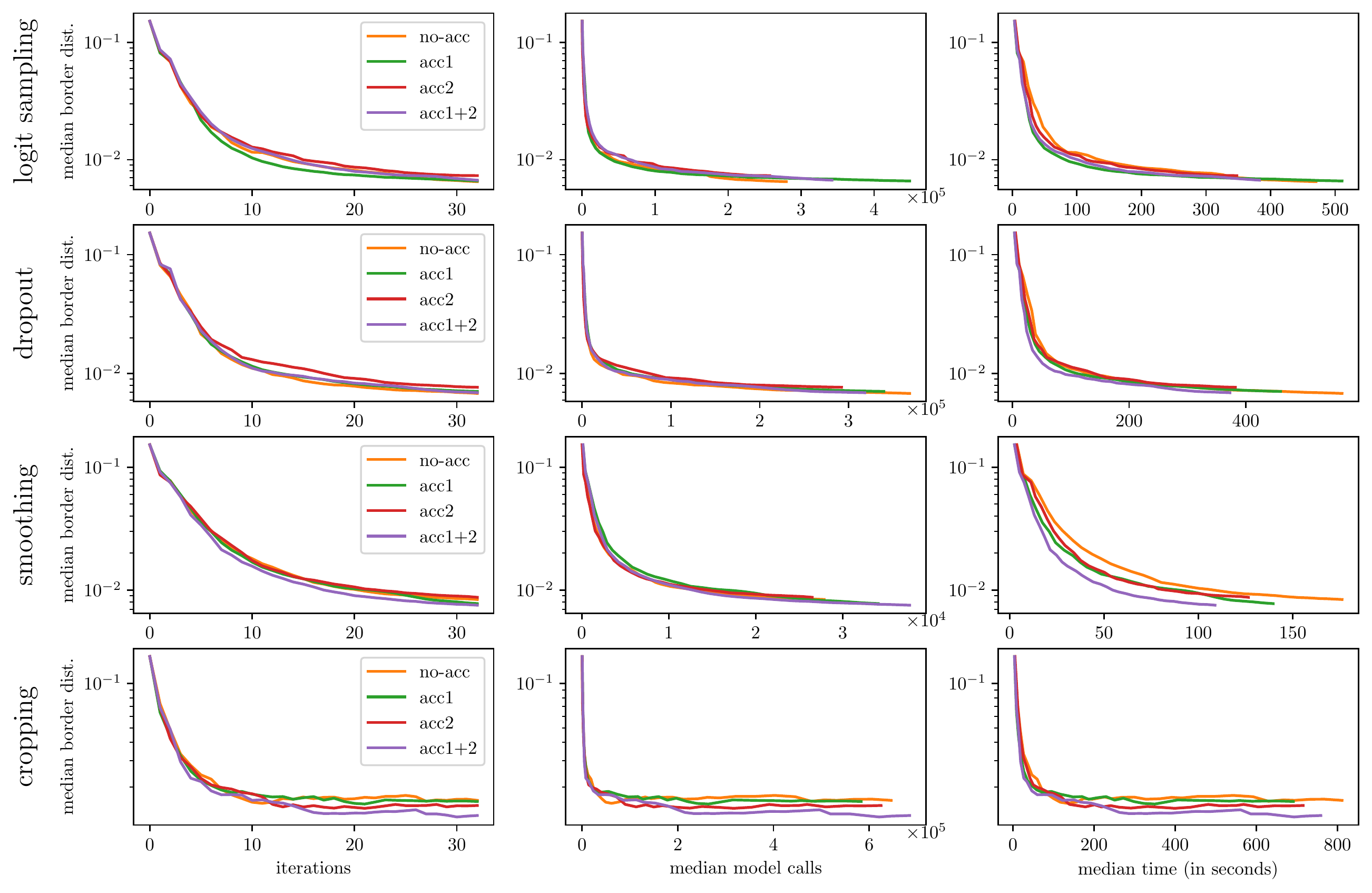}
    \caption{\textbf{CIFAR10}: Same comments as for \cref{fig:acceleration_mnist}.}
    \label{fig:acceleration_cifar}%
\end{figure}

We tested these acceleration tricks on 20 images with results shown in \cref{fig:timing_acceleration,fig:queries_acceleration,fig:acceleration_mnist,fig:acceleration_cifar}.
\Cref{fig:timing_acceleration} confirms that both tricks accelerate the bin-search steps and can be combined for further acceleration.
\Cref{fig:queries_acceleration} confirms that trick~\ref{it:acc1}, despite decreasing wall-clock time, increases the amount of bin-search queries, and that trick~\ref{it:acc2} decreases it (with tighter priors we need less queries to determine the parameters up to a given precision).
\Cref{fig:acceleration_mnist,fig:acceleration_cifar} show that, despite accelerating the attack (column c and \cref{fig:timing_acceleration}), the tricks do not significantly affect PSJ's output quality (column a) and number of median model queries (column c).

\end{document}